# Towards a Holistic Framework for Multimodal Large Language Models in Three-dimensional Brain CT Report Generation

Cheng-Yi Li[1,2,3], Kao-Jung Chang[2,4,5,6]#, Cheng-Fu Yang[1], Hsin-Yu Wu[2,3], Wenting Chen[7], Hritik Bansal[1], Ling Chen[8], Yi-Ping Yang[2], Yu-Chun Chen[3,5,8,9], Shih-Pin Chen[4,10], Jiing-Feng Lirng[11,12], Kai-Wei Chang[1]#, Shih-Hwa Chiou[2,4,6]#

[1] Department of Computer Science, University of California, Los Angeles
[2] Department of Medical Research, Taipei Veterans General Hospital
[3] School of Medicine, College of Medicine, National Yang Ming Chiao Tung University
[4] Institute of Clinical Medicine, National Yang Ming Chiao Tung University
[5] Big Data Center, Taipei Veterans General Hospital
[6] Department of Ophthalmology, Taipei Veterans General Hospital
[7] Department of Electrical Engineering, City University of Hong Kong
[8] Institute of Hospital and Health Care Administration, National Yang Ming Chiao Tung University
[9] Department of Family Medicine, Taipei Veterans General Hospital
[10] Department of Neurology, Neurological Institute, Taipei Veterans General Hospital
[11] Department of Radiology, School of Medicine, National Yang Ming Chiao Tung University
[12] College of Medicine, National Yang Ming Chiao Tung University

# These authors share equal correspondence responsibility.

*Abstract* — Multi-modal large language models (MLLMs) have been given free rein to explore exciting medical applications with a primary focus on radiology report generation. Nevertheless, the preliminary MLLM successful attempts in 2D medical image-text pair captioning are incompetent to reflect the real-world diagnostic challenge in the volumetric 3D anatomy. Toward deploying MLLM for more applicable diagnostic context, we noticed that the (1) scarcity of 3D image training dataset, (2) the direct use of undifferentiated foundation MLLMs, and (3) the lack of pertinent caption evaluation metrics were independent domain-specific constraints that integratively hobbles the iteration of next-generation medical MLLM research. In this regard, this study collected a 3D-BrainCT dataset (18,885 text-scan pairs) and applied clinical visual instruction tuning (CVIT) to train volumetric anatomy-sensible BrainGPT models to generate radiology-adherent 3D brain CT reports. Statistically, our BrainGPT model scored BLEU-1 = 44.35, BLEU-4 = 20.38, METEOR = 30.13, ROUGE-L = 47.6, and CIDEr-R = 211.77 during internal testing and demonstrated an accuracy of 0.91 in captioning midline shifts on the external validation CQ500 dataset. By further inspecting the captioned report, we reported that the traditional metrics appeared to measure only the surface text similarity and failed to gauge the information density of the diagnostic purpose. To close this gap, we proposed a novel Feature-Oriented Radiology Task Evaluation (FORTE) to estimate the clinical relevance (lesion feature and landmarks) of the report. Notably, the BrainGPT model scored an average FORTE 0.71 F1-score (degree=0.661; landmark=0.706; feature=0.693, and impression=0.779). To demonstrate that BrainGPT models possess objective readiness to generate human-like radiology reports, we conducted a Turing test that enrolled 11 physician evaluators, and around 74% of the BrainGPT-generated captions were indistinguishable from those written by humans. While various computational intelligence researchers have advocated the avant-garde MLLM applications, our work embodies a holistic framework that showcased the first-hand experience of curating a 3D brain CT dataset, fine-tuning anatomy-sensible language models, and proposing robust radiology evaluation metrics. We deemed that the adventure of docking MLLM for 3D brain CT report generation may unfold new MLLM applications at the forefront of human-machine collaborated modern healthcare.

## Introduction

The rapid ascent of artificial intelligence (AI) technologies has profoundly influenced the interpretation of multimodal information such as computed tomography (CT),[1] gross dermatologic conditions,[2] cytology slides,[3] ophthalmology images,[4] etc. While modern models excel in tasks like classification, regression, and segmentation, they fall short in providing the natural-language expressivity and interactivity required in hands-on clinical workflows.[5] This discrepancy is critical because instant scan interpretation is essential to clinical decision-making, with the rate decision step being the radiologist's report writing.[6,7] To address this need, early generative models.[8-11] have been applied to chest X-ray (CXR) report generation in datasets such as the MIMIC-CXR and OpenI to relieve radiologists' time-consuming and tedious burden.[12,13] These studies demonstrated a potential interface where human-computer collaboration can form medical reports, which can be further refined with multi-modal large language models (MLLM).[14]

To examine the full adequacy of MLLM in radiology report generation, we conducted a thorough review of relative works. Despite the primary success in CXR report generation, we identified three crucial limitations in the existing literature, including (1) data complexity, (2) model capacity, and (3) evaluation metric fidelity. These innate limitations have hindered the development of totipotent MLLMs.

While the human body is inherently three-dimensional (3D), existing 2D CXR image datasets[12,13] have yet to encompass (1) pathologic features in complex neuro-vascular anatomy, i.e., the brain, heart, and eye, (2) spatial landmarks within the 3D image context, and (3) the intensity degree or size of lesions. Moreover, the single-slice image content from current CT image-text data is usually selected from 3D data to include the described feature. This may be subject to *sharpshooter fallacy,* where the selected slice often indicates the caption target of interest. Hence, it is imperative to curate a 3D image set to inspect the dexterities of MLLMs in real-world diagnostic contexts.

Crucially, MLLMs must be capable of multi-image reasoning or video understanding when used to generate reports for real-world 3D data. Recent medical MLLM studies such as LLaVA-Med (Microsoft) and Med-PaLM M (Google Research) have demonstrated promising X-ray and single-slice CT report generation performance.[15,16] However, during testing, the brain CT data is presented in a 3D format, which poses challenges for the MLLM to identify the specific slice containing the lesion of interest precisely *in situ*.

3D CT scans of the brain, with reports structured in a list-by-list format focusing on differential diagnoses, serve as the first-line diagnostic for diverse intracranial conditions.[17,18] The brain lesion's degree, size, and location are paramount to forming precise diagnoses and guiding subsequent clinical decisions. Therefore, the value of key radiology descriptors outweighs grammatical filler words. In this regard, traditional evaluation metrics, designed to evaluate short translation[19,20], abstract summary tasks[21], and common image captioning[22] cannot gauge the clinical essence of Brain CT reports.[23] On the other hand, the CheXpert clinical efficacy evaluation is designed to capture pathology keyword retrieval rates to represent the CXR report quality.[11,24] However, 14-feature retrieval rates cannot reflect the multi-semantic context (inclusive or exclusive; progression or resolution; relative location and size) of disease features. Moreover, the uncategorized keyword bank concept may not be transferable from one image modality to other modalities. To alleviate such constraints, radiology semantic categories must be incorporated into a novel scoring system to measure the clinical essence of radiology reports.

In this study, we aim to advance the application of MLLMs in radiology by addressing several critical areas:

(1) We curated a large-scale 3D-brainCT dataset (18,885 text-scan pairs) abundant with lesion degree, spatial landmarks, and diagnostic impressions of neuronal and vascular features to present a real-world challenge for MLLMs. (Fig. 1 and Extended Data Table 1)

(2) We applied the clinical visual instruction tuning (CVIT) concept to train the BrainGPT model from the open-source Otter.[25,26] The Otter model's multi-image captioning capacity and clinical instruction design enable the generation of clinically sensible captions from volumetric brain CT scans. (Fig. 2) The BrainGPT was externally validated on CQ500 (Fig. 5) and an 11-physician rater Turing Test-like linguistic-embedded evaluation. (Fig. 6)

(3) We proposed feature-oriented radiology task evaluation (FORTE) to measure the clinical relevance of captions based on distinct disease objectives, including 4 generalizable keyword categories (degree, landmark, feature, and impression). (Fig. 4 and Extended Data Fig. 2) We also suggested that preprocessing processes, such as sentence pairing and negation removal, are beneficial in adapting traditional metrics to the detailed, itemized, and differential diagnosis-oriented format of brain CT reports. (Fig. 3 and 4c)

Overall, this study proposes a holistic framework that addresses the implementation details of applying MLLMs in brain CT and other complex anatomy images.

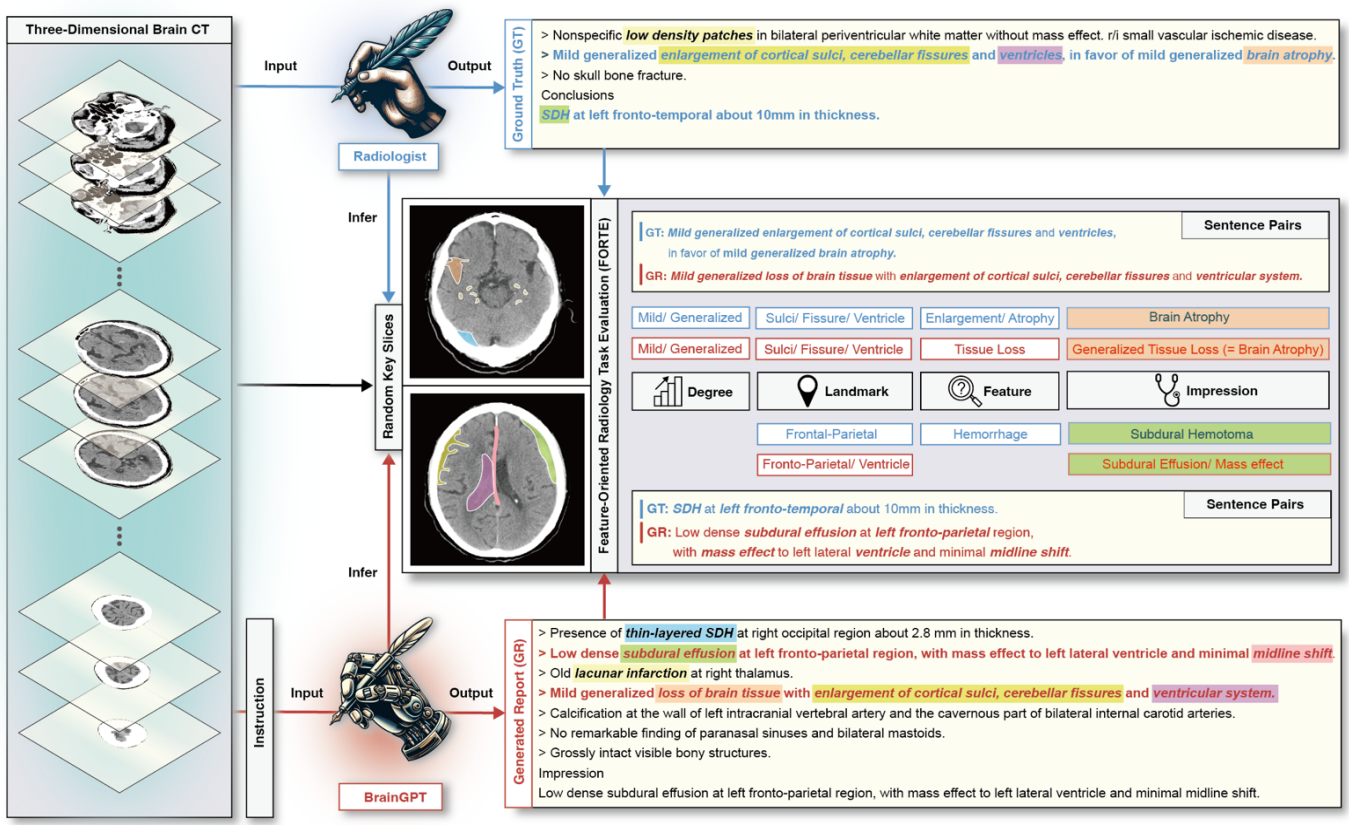

**Fig. 1 | Study Overview.** Given a 3D brain CT scan, BrainGPT can follow instructions to generate an image report for the volumetric data. We then evaluate the generated report in comparison with the ground truth by radiologists. Within the proposed feature-oriented radiology task evaluation (FORTE), reports are reassembled into four categories of clinical keywords: degree, landmark, feature, and impression. The system accommodates synonyms, allowing the recognition of a broader array of related terms that may appear in different contexts to address the challenge of lexical variability inherent in clinical reports. In this image-text example, image features and corresponding report descriptions are labeled with the same color.

## Results

### Training BrainGPT with Clinical Visual Instruction Tuning

To implement medical MLLM in 3D brain CT report generation, we collected a real-world 3D-BrainCT dataset and trained BrainGPT models by applying visual instruction tuning on the Otter foundation model (Extended Data Fig. 4). Our fine-tuning conditions included regular visual instruction tuning (RVIT): (1) Plain Instruction (conveys the model's role as a radiology assistant) and (2) In-context example Instruction[27] (3-shot examples added to the Plain Instruction), or clinical visual instruction tuning (CVIT): (3) Template Instruction (structured clinical-defined QA templates added to the Plain Instruction) and (4) Keyword Instruction (categorical guidelines focused on keywords added to the Plain Instruction). As a result, we derived four BrainGPT models (-plain, -example, -template, -keyword) that are tuned to perform CT interpretation with different prior levels of clinical sensibility (Fig. 2).[25]

At the whole report level, all of the four fine-tuned BrainGPT models outperformed the baseline Otter model in traditional metrics (Mann-Whitney U test $p < 0.01$, see Fig. 3b). In particular, the baseline Otter model scored the lowest in BLEU-4 (0) and CIDEr-R (5.9) metrics, indicating that the baseline model-generated CT report had low n-gram overlap and underfit semantic term usage frequency. In addition, we observed a mismatch between the conventional metric and the report quality by reviewing generated reports (see Extended Data Table 1). Moreover, no traditional metrics, except CIDEr-R, rated the BrainGPT performance in accord with the crescendo elaborateness of RVIT and CVIT (Mann-Whitney U test $p > 0.05$ across four visual instruction tuning conditions, summarized in Extended Data Table 2). Therefore, we demonstrated that traditional metrics were inherently insensitive to the clinical essence of the generated radiology reports.

### Sentence pairing

To investigate whether the list-by-list report architecture of brain CT images accounted for low traditional metric scores, we applied sentence pairing to decompose the multi-sentence paragraph into smaller semantic granularity. (Fig. 3a) As a result, the process of sentence pairing increased traditional metric scores by an average of 5.28 points in METEOR, 6.48 points in ROUGE-L, and an astonishing 114 points in CIDEr-R (Fig. 3b and Extended Data Table 2). The traditional metric scores for every BrainGPT model were significantly improved across most traditional metrics ($p < 0.001$, as shown in Fig.3c and Extended Data Table 3). In addition, the score distribution indicated significant linear correlation (Pearson correlation coefficient $r > 0.7$) in BLEU-2, BLEU-3, BLEU-4, METEOR, and ROUGE-L with statistical significance ($p$-value $< 0.05$). However, apart from the other n-gram-based traditional metric, the CIDEr-R score gained a more positive shift from the process of sentence pairing ($r = 0.577$).

Moreover, we observed a positive correlation between advanced CVIT and increased traditional metric scores. In particular, the CIDEr-R score depicted the most prominent ascending trend across the hierarchically fine-tuned BrainGPT-plain (125.86), BrainGPT-example (132.38), BrainGPT-template (147.92), and BrainGPT-keyword (153.3). This sentence pairing procedure not only relieved the sequential constraint between the report input and generated output but also accentuated the clinical value in the list-by-list differential diagnosis reports.

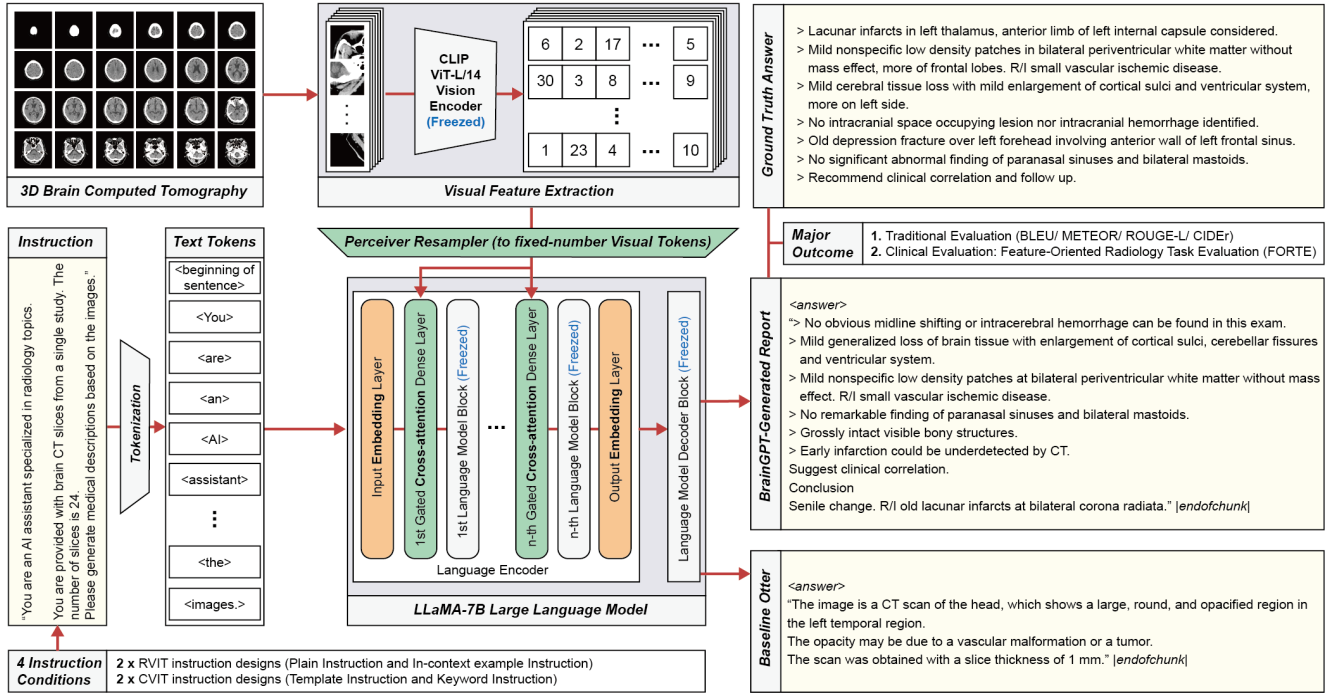

**Fig. 2 | Clinical visual instruction tuning BrainGPT from Otter.** Our approach adopts the end-to-end Otter paradigm to train the BrainGPT model. We designed four distinct fine-tuning conditions: two for regular visual instruction tuning (RVIT) – Plain Instruction and In-context Example Instruction, and two for clinical visual instruction tuning (CVIT) – Template Instruction and Keyword Instruction. To enable multi-image in-context learning, we formatted the data as image-instruction-answer triplets, where instructions were tokenized, and images were enhanced before being inputted into the model. The Otter framework integrates visual data (using a frozen CLIP ViT-L/14 encoder) and language data (using the LLaMA-7B large language model) through a trainable perceiver resampler module. In the LLaMA-7B architecture, cross-gated attention layers were added to focus on CT slices, while all other modules, except for input/output embeddings, were frozen to minimize training expenses. The fine-tuned models were evaluated using traditional metrics and a novel FORTE evaluation method, designed for this study.

**Feature-Oriented Radiology Task Evaluation (FORTE)**

Given that CIDEr-R captures the hierarchical clinical essence across visual instruction tuning conditions, we hypothesized that its frequency-inverse document frequency (TF-IDF) component reacts to the usage of rare radiology keywords in reports. Therefore, we analyzed the term frequency in both the ground truth and test outputs (Fig. 4$a_1$). We found that the baseline Otter model had a low recall rate for keywords, whereas the BrainGPT models, after visual instruction tuning, demonstrated significant radiology keyword usage in image captioning. Notably, the BrainGPT-keyword (Fig. 4$a_2$: y=0.95x-1.18; $r^2$=0.85) maintained a high recall rate across varying frequencies of radiology keywords.

Due to the potential instability caused by keyword frequency variance in report evaluations, we introduced the feature-oriented radiology task evaluation (FORTE). This method gauges the medical content of the generated report by focusing on the density of radiology information. We categorized radiology keywords into degree, landmark, feature, and impression subsets to offer a multi-faceted evaluation of the system's performance.

By evaluating the F1 score of each FORTE-measured keyword category (see Extended Data Table 4), we noted that the advanced CVIT models, BrainGPT-template and BrainGPT-keyword, scored higher average F1 score than the RVIT models. (Mann-Whitney U test $p$-value <0.001, as detailed in Extended Data Table 5) Specifically, BrainGPT-keyword captured the most radiology details, with FORTE F1 scores of: degree=0.548, landmark=0.533, feature=0.574, and impression=0.649. These collective results indicate that clinical instruction-tuned BrainGPT models generate brain CT reports with a higher level of radiology term usage and better alignment with original diagnostic reports.

**Compare FORTE against traditional evaluation metrics**

To determine if FORTE reveals report qualities uncovered by traditional metrics, we conducted a Pearson correlation coefficient analysis to compare FORTE with traditional metrics (Fig. 4d). We observed that traditional metrics had high intra-correlations (r > 0.7, $p < 0.001$), but lower correlation with FORTE (r < 0.5, $p < 0.001$) with the FORTE. In addition, the four FORTE domains (degree, landmark, feature, and impression) had lower intra-correlations (r < 0.5, $p < 0.001$ compared to traditional metrics). This comparison indicated that the FORTE addresses broader and distinct disease aspects than those traditional metrics depicted.

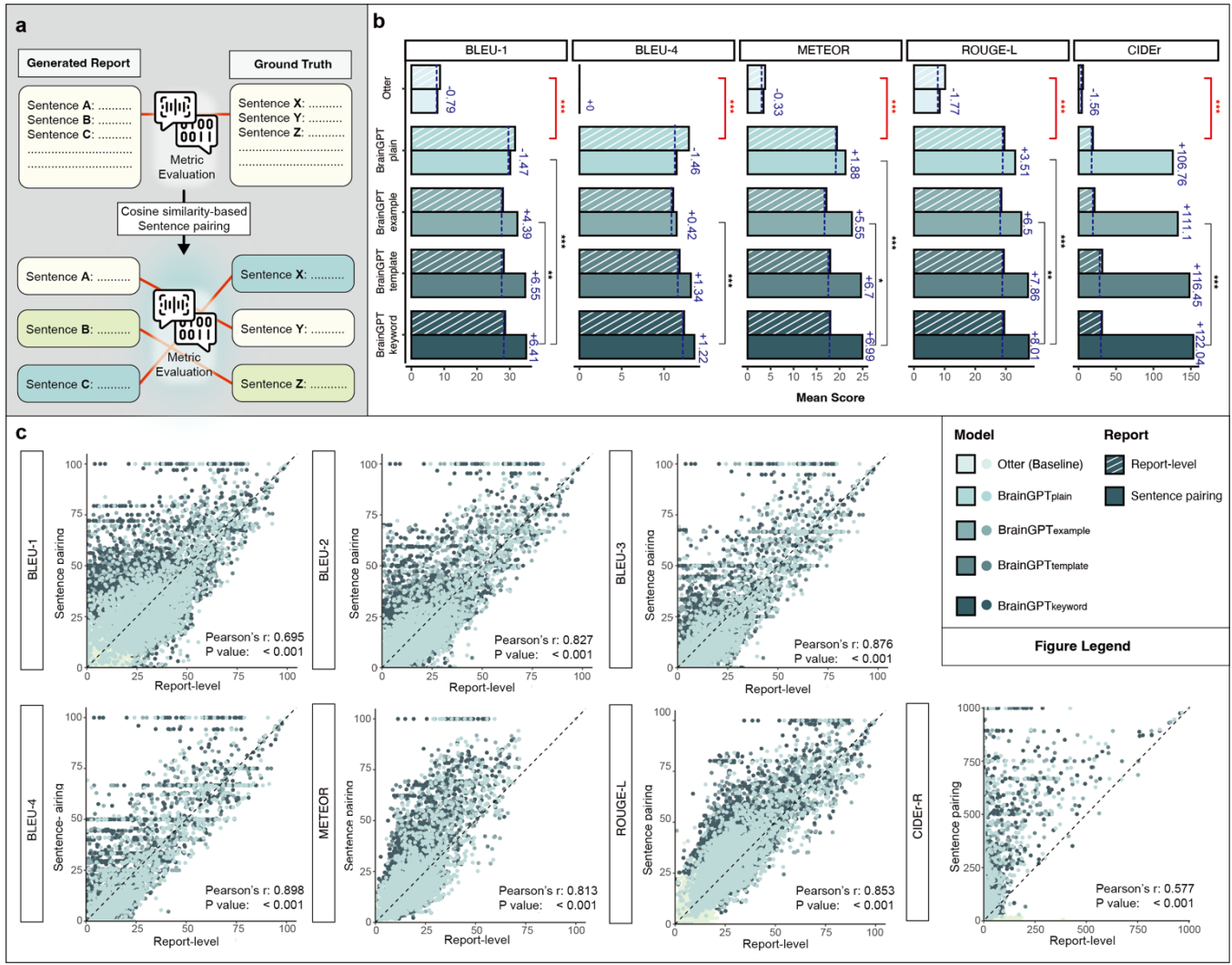

**Fig. 3 | Improving conventional captioning metrics.** We performed a traditional evaluation comparing the BrainGPT report and the radiologist's report. (a) The generated report and the ground truth, each containing multiple descriptions, were restructured with sentence pairing to simulate a common image captioning scenario. (b) Traditional metric evaluation results show the training efficacy with statistically significant differences between the Otter model and the fine-tuned BrainGPT models (indicated by red annotations). The preprocessing involving sentence pairing highlighted the clinical value in the list-by-list differential diagnosis reports, as the CVIT (Template Instruction and Keyword Instructions) achieved significantly better results. (c) The dashed line denotes equivalence, and each data point corresponds to a text-image instance. The distribution of sentence pairing effects demonstrates the unique status of CIDEr, which is the only metric showing a moderate Pearson's r. (* $p < 0.05$, ** $p < 0.01$, *** $p < 0.001$)

**Negation Removal: Effect of "Interpretation Spree"**

Upon further review of keyword usage frequency, we found that "no" was notably among the most frequently used words, along with "of," "and," "in," and "the." This contradicts the common belief of "reporting bias," where it is assumed that language models, similar to laypersons, are less likely to mention negative features. Clinically, radiologists often focus on context-oriented descriptions to "rule out" diagnostic targets, a practice not mirrored by BrainGPT. Consequently, BrainGPT's reports were overlaid with excessive negative descriptions rather than the concise, context-specific language used by radiologists, a phenomenon we termed the "Interpretation spree."

To address this, we applied negation removal to limit descriptions to positive findings. (Fig. 4c) The average score increases were: BLEU-1= 29.25%, BLEU-2= 34.9%, BLEU3= 50.54%, BLEU-4= 57.26%, METEOR= 24.97%, ROUGE_L= 29.04%, CIDEr-R =

46.6%, with statistical significance (See Extended Data Fig. 1).

Additionally, negation removal improved the BrainGPT FORTE keyword average F1 score by 0.118 for degree, 0.194 for landmark, 0.139 for feature, and 0.163 for impression (see Fig. 4b and Extended Data Table 4). Negation removal led to an overall average F1 score gain of 0.153, with the best instruction-tuned model, BrainGPT-keyword, seeing its F1 score rise from 0.576 to 0.71. Together, we demonstrated that negation removal allowed evaluation metrics to focus on positive findings. Meanwhile, negation removal also enhanced average traditional scores and FORTE F1 scores.

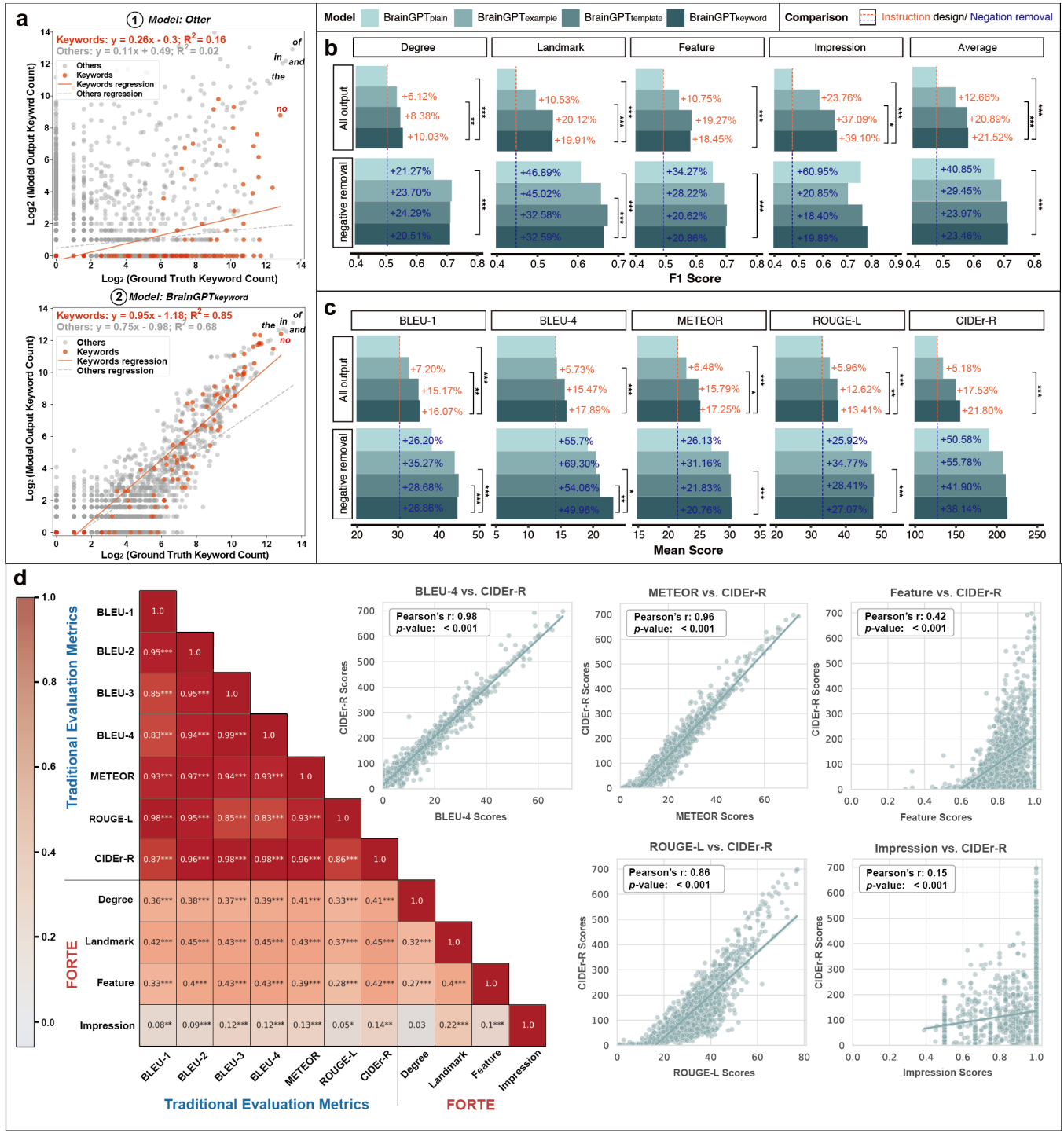

**Fig. 4 | Feature-Oriented Radiology Task Evaluation (FORTE)** (a) Distribution of keyword frequency in the ground truth descriptions versus the BrainGPT model outputs. Most high-frequency keywords are grammatical filler words, but the importance of the keyword "no" is highlighted. (b) Results for FORTE, with the four categories designed to measure the clinical essence of generated reports. Negation removal was implemented to limit the description range to only positive findings. (c) Effects of applying negation removal on traditional evaluation metric results. (d) Pearson analysis between the traditional metrics and the FORTE keyword categories. (* $p < 0.05$, ** $p < 0.01$, *** $p < 0.001$)

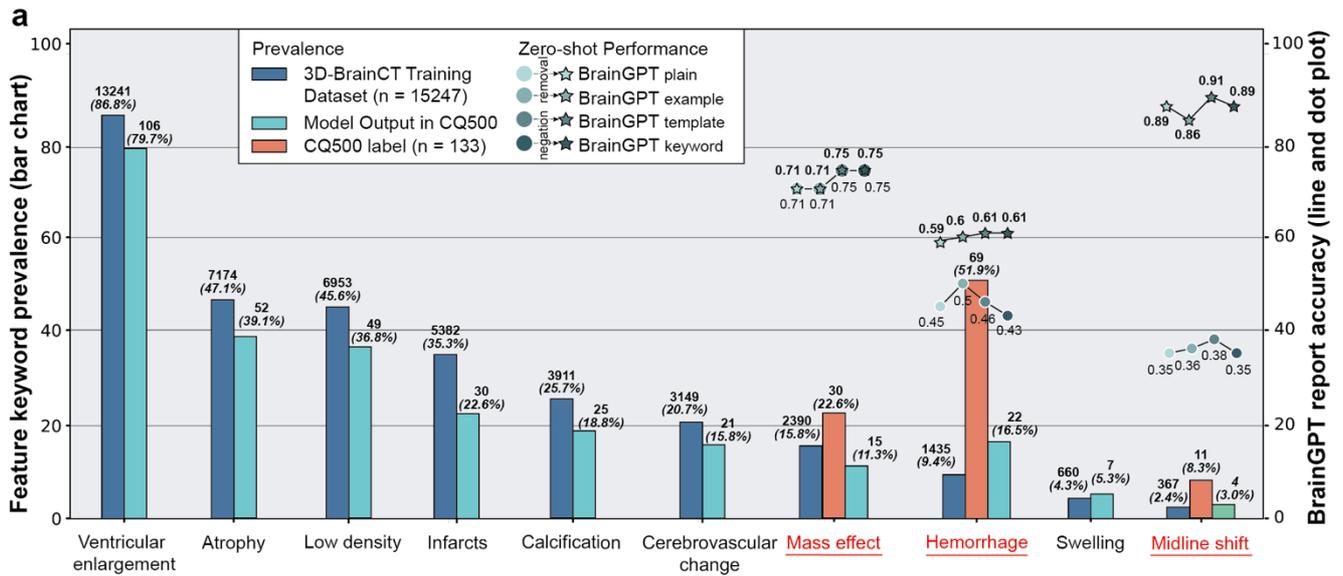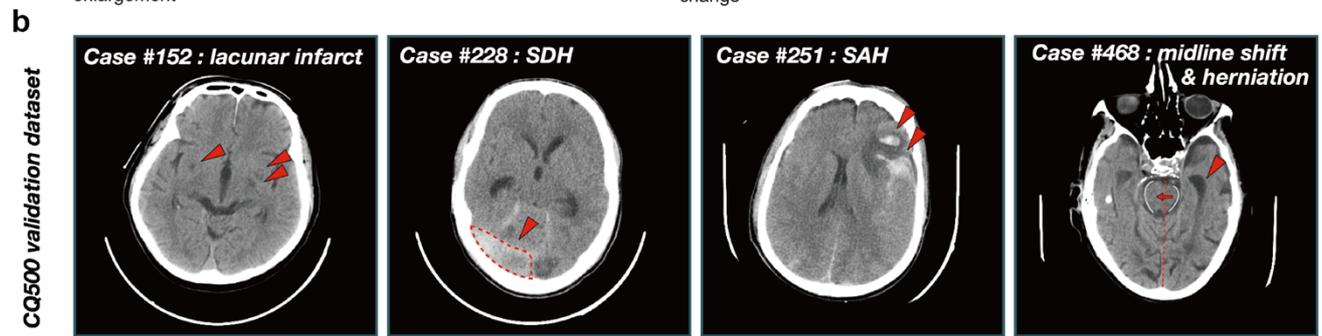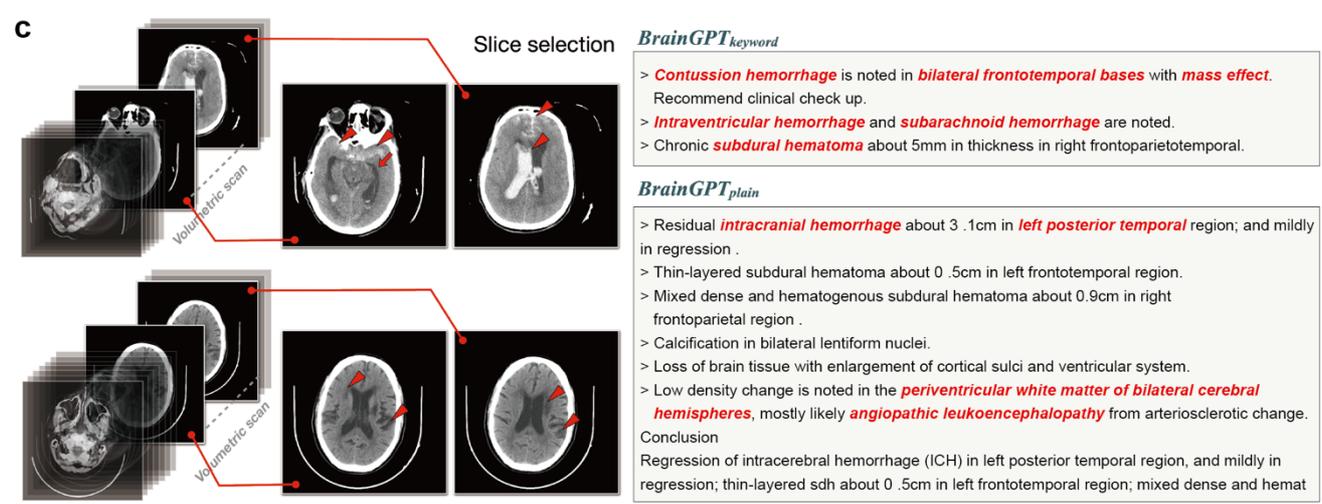

**Fig. 5 | Examining Multi-object Multi-Slice Captioning Capability of BrainGPT on the Stroke-featured CQ500 External Validation dataset.** (a) Bar chart summarizing and comparing the feature composition between our 3D-brainCT (training and internal testing) dataset and the CQ500 (external validation) dataset. The feature captioning accuracy (mass effect, hemorrhage, and midline shift) of the four BrainGPT models was calculated and represented as a line plot. (b) Head-to-head comparison of BrainGPT model-generated image reports for distinct brain lesion features (lacunar infarction, subdural hemorrhage (SDH), subarachnoid hemorrhage (SAH), midline shift, and herniation). (c) BrainGPT demonstrated a strong captioning ability for volumetric brain CT slices, generating reports with lesion descriptions linked to individual image slices.

### Assess BrainGPT Generalization by CQ500 External Validation

We conducted zero-shot external validation on the CQ500 brain CT dataset (n=133) to assess BrainGPT's ability to perform report captioning in generalized clinical conditions. We estimated the clinical keyword retrieval rate by comparing BrainGPT's reports to the ground truth labels of the CQ500 CT scans (Fig. 5a). Notably, BrainGPT mentioned keywords (e.g., ventricular enlargement, atrophy, infarction, and mass effect) in the generated reports for the CQ500 validation dataset with frequencies similar to the training dataset (3D-brainCT). These zero-shot captioning results indicated that BrainGPT had inscribed the CT report phraseology and writing structure for brain disease differential diagnosis.

CQ500 is an intracranial hemorrhage (ICH) dataset encompassing "mass effect," "hemorrhage," and "midline shift" features, which were less indexed in our geriatric population 3D-brainCT dataset. Nevertheless, the four instruction-tuned BrainGPT models achieved fair accuracy in reporting mass effect (0.71-0.75), midline shift (0.35-0.38), and hemorrhagic events (0.43-0.5). Furthermore, consistent with our previous findings, negation removal (aligning with differential diagnosis interest) significantly increased report accuracy for midline shift (0.86-0.91) and hemorrhagic events (0.59-0.61). Notably, in terms of diagnosing "mass effect," "hemorrhage," and "midline shift," the advanced CVIT models (BrainGPT-template and BrainGPT-keyword) generally outperformed the RVIT fine-tuned models (BrainGPT-plain and BrainGPT-example). This suggests that CVIT instructions acted as a pre-model rectifier, enhancing the focus on valuable clinical etiologies.

To further investigate diagnostic credibility, we examined report results for lacunar infarct, subdural hemorrhage (SDH), subarachnoid hemorrhage (SAH), and midline shift with mass effect (Fig. 5b). Interestingly, in the lacunar infarction case, BrainGPT-example accurately identified the hypodense old lacunar infarct in the putamen and thalamus regions. However, "putamen" was misspelled as "putmen," likely due to a typo learned from the original training report. The model also generated successful captions for brain lesions, including SDH at the right occipital, SAH, and left uncal herniation. To examine if BrainGPT was subject to the *sharpshooter fallacy* in 3D CT scans, we isolated multi-object cases with different lesion features in various CT cuts (Fig. 5c). We noted that in some multi-lesional brain CT, the BrainGPT reports could recognize distinct hemorrhage subtypes (contusion, intraventricular and subarachnoid hemorrhage) with a sparse comprehension of the lesion location (frontotemporal lobe). In another case, BrainGPT described a regressing chronic ICH and a periventricular angiopathic leukoencephalopathy with white matter hypo-density. These multi-object, multi-slice caption reports demonstrated BrainGPT's potential to provide differential diagnoses for hemorrhages and hypodense image features.

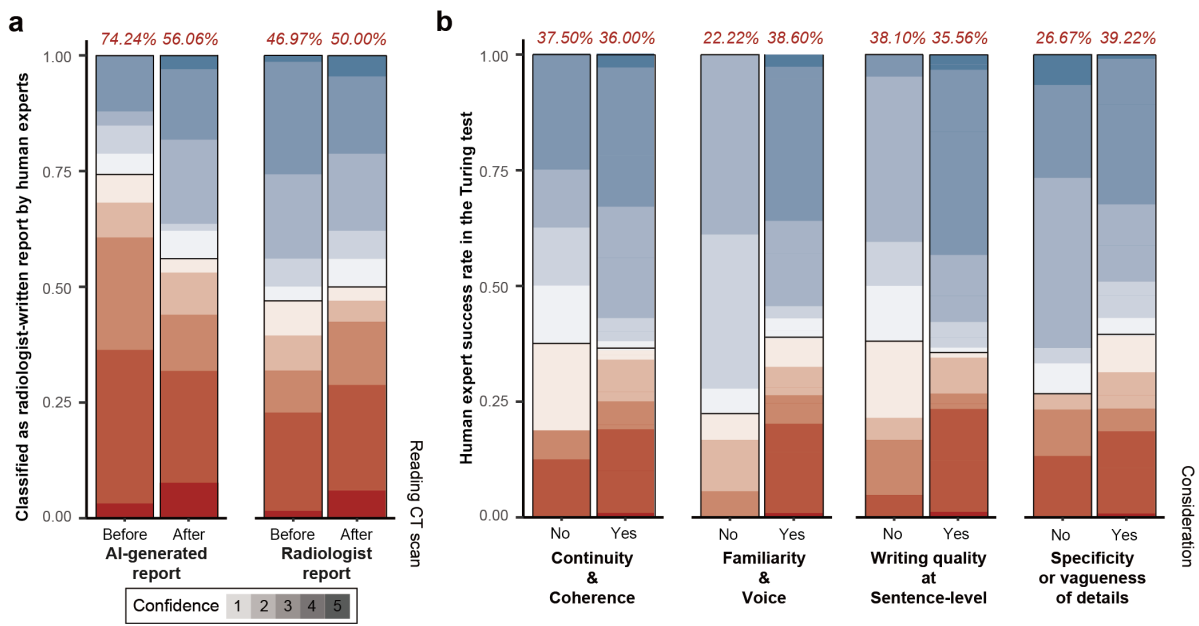

**Fig. 6 | *Turing* test.** (a) the accuracy of responses and levels of confidence among physicians before and after review of the original CT scans. (b) the linguistic criteria employed by physicians and the corresponding accuracy change.

**Linguistic-embedded *Turing* test**

Our results demonstrated that the BrainGPT models were capable of emulating the writing style of radiologists, using domain-specific language, and providing preliminary diagnostic information. To further evaluate whether these BrainGPT-generated reports could be distinguished by healthcare providers, we conducted a *Turing* test. (For the questionnaire design, see Extended Data Fig. 5)

In this test, we enrolled 11 physicians (2 radiologists, 2 neurologists, and 7 other licensed medical doctors) (Fig. 6a). These doctors were asked to determine whether 6 provided report cases (two lacunar infarcts, cortex atrophy, subdural hemorrhage, and midline shift) were written by machines or humans. Interestingly, out of 66 evaluations, 74.24% of BrainGPT-generated captions were mistakenly identified as human-written, whereas only 46.97% of human-written reports were correctly identified. When we provided the evaluators with the input CT images, nearly 18% of their decisions were changed for BrainGPT reports, resulting in 56% of machine-written reports being indistinguishable from human-written reports. This input image-directed decision change was not observed in real human-written reports, which had a 50% correct identification rate. Overall, integrating the scan-text context into the *Turing* test led to more confident evaluations on the 5-point Likert scale ($p$-value). This suggests that reviewing actual image contexts helps human evaluators detect subtle differences between machine and human-written reports, likely due to off-target descriptions and inadequate precision in machine-generated captions.

We also analyzed the qualitative traits influencing evaluator decisions (Fig. 6b). We found that the traits of "Familiarity and voice" (29%), "Specificity or vagueness of details" (37.60%), "Continuity and Coherence" (33.33%), and "Writing quality at sentence level" (33.33%) were significant factors in the evaluators' decisions.[28] Notably, when reviewers considered "Familiarity and voice," the average success rate increased by 73.7% (from 22.22% to 38.6%). Similarly, considering "Specificity or vagueness of details" increased the average success rate by 47% (from 26.67% to 39.22%). These linguistic criteria underscore the importance of fine-tuning the radiologist writing style and incorporating radiology-specific details in medical report generation.

## Discussion

In this study, we assembled a holistic framework to address various unmet needs in achieving robust report generation for 3D brain CT. Based on our collective experiences, we recommend applying the following sequential protocol (or an equivalent process) to ensure caption and evaluation rigor:

1. CVIT: Incorporate clinical domain knowledge in implementing visual instruction tuning

2. Combine "sentence pairing" and "negation removal": Adapt traditional metrics to radiology reports

3. Employ FORTE: Gauge the clinical essence of caption results in multiple aspects

4. Conduct linguistic-embedded *Turing* test: Include the corresponding image and linguistic criteria in human expert evaluation

Based on these technical approaches, we demonstrated that the fine-tuned open-source MLLMs can be instrumented for 3D medical image interpretation.

Recent work by Hamamci et al. showed that generative models are competent in 3D chest CT report generation at a state-of-the-art (SOTA) level (BLEU-1 = 46, BLEU-4 = 36.9, METEOR = 29.5, ROUGE-L = 45.9).[29-31] However, their customized transformer model required 7 days of training on a single NVIDIA A100 GPU, whereas our BrainGPT requires just 12 hours of fine-tuning on two NVIDIA A100 GPUs. Additionally, Google AI's Med-Gemini-3D can perform 3D CT report generation, but only 53% of its reports were considered clinically valid in human evaluations.[32] The high computational cost of using large-scale Google TPUv4 accelerator pods makes this approach infeasible for general research with limited resources. In contrast, our BrainGPT uses an end-to-end open-source Otter framework (CLIP ViT-L/14 vision encoder[32] and LLaMA-7B[33]), which allows for experiment replication and checkpoint sharing. Moreover, the reduced training cost of BrainGPT enables efficient visual instruction tuning, enhancing model performance and tailoring responses to specialized or stylistic requirements (Extended Data Table 6).

While we did not modify the Otter model structure, we attribute the SOTA-level performance to the combined effects of RVIT and CVIT. Singhal et al. first explored task-specific RVIT in the medical domain, showing that chatbot performance improved with medical QA in-context example primings.[27] Similarly, Med-PaLM M used image cues (CXR and pathology slides) alongside clinical instructions to guide MLLM on multimodal medical tasks.[16] Echoing these studies, our CVIT models (BrainGPT-template, BrainGPT-keyword) outperformed RVIT models in brain CT captioning. This suggests that fine-grained, specialist-grade instruction design may optimize model outcomes for clinical captioning tasks.

We also highlighted that traditional metrics are unfit for evaluating clinical captioning tasks. Medical image reports assist in differential diagnosis and thus are characterized by complex paraphrasing, high token counts (>100), and numerous negative descriptions, which conflict with common metric evaluation contexts.[23,34] Additionally, we observed an "Interpretation Spree" behavior, where BrainGPT provided off-target (but not hallucinated) diagnostic narrations from multi-object brain CT contexts (detailed in Supplementary Table 1). This behavior is detrimental because (1) off-targeting may exclude the primary disease focus (e.g., stroke or brain tumor), and (2) expanded narration may dilute traditional metrics, leading to invalid evaluations.

To address low traditional metric values from different narration aims, we preprocessed radiology reports with "Sentence Pairing" and "Negation Removal." We then applied FORTE to decompose BrainGPT reports into keyword categories addressing effective radiology descriptor frequency. The application gap is attributed to surface-form similarity and lexicographic analytic restrictions.[34] For the first time, we performed Pearson correlation analysis on different evaluation metrics, finding that FORTE encapsulates the semantic dimension more effectively than traditional CE metrics.[24] Meanwhile, the

FORTE category concept is customizable and transferable across various medical image captioning tasks by substituting categorical keyword bank contents. (Extended Data Fig. 2)

Human expert evaluation in natural language processing experiments has been conducted under distinct experiment designs and served diverse study purposes. Therefore, the resulting perspectives were often inconsistent and non-comparable across individual contexts. To this end, related works had instrumented quantitative (completeness, correctness, conciseness) and qualitative (content, language, structure) measurements to dissect the riveting characters that discern synthesized clinical reports from human reports.[35-37] By adopting similar designs with objective linguistic criteria, we found that both reviewer success rate and answer alternation reasons ("Suspicious wording" and "Both not mentioning critical features") were associated with the writing style ("Familiarity and voice" and "Specificity or vagueness of details") rather than the sentence-level writing quality and coherence.[28] The significance of medical report writing style was also underlined in an independent prompting study.[36] Interestingly, we observed that input case imbalance can influence caption writing style, potentially related to over-fitting noticed in general model training (see also Extended Data Fig. 3).

This study has several limitations, which should be addressed in future works. First, this pilot brain CT study has no counterpart MLLM module to benchmark, and hence we cannot justify the efficacy at a SOTA level; however, we applied external validation to ensure the caption validity in the brain CT module. Second, BrainGPT is trained on degeneration-oriented data and thus it fails to caption the malignancy tumor and acute traumatic features in CQ500. This phenomenon reflects that the training material may prime the dexterity of the resultant module.[38] Hence, we suggest enrolling diverse disease etiologies with the aim of differential diagnosis to improve MLLM generalization on border brain CT features. Last, we conducted CVIT and invented clinical-oriented evaluations (sentence-pairing, negation removal, and FORTE), but we did not experiment on whether changing the model backbone could benefit brain CT captioning. Future research avenues could be comparing multi-model results and fine-tuning the vision encoder and the language model for CT.

clinical knowledge. *Nature* **620**, 172-180 (2023).

## Method

### Study Design and Oversight

In this study, we trained BrainGPT to generate 3D brain CT reports. Then, we examined the caption efficacy by (1) adapting traditional evaluation metrics (2) proposing a clinical-oriented FORTE metric, (3) applying an external validation CQ500, and (4) conducting the linguistic-embedded *Turing* test.

### Study Patient

We collected 18,885 brain CT scans (742,501 slices) from 9,689 patients with Alzheimer's Disease (mean age = 82.59 years [standard deviation = 9.3 years]; 56.4% male) at Taipei Veterans General Hospital in Taipei, Taiwan, between January 1, 2010, and December 31, 2022. All data were collected under institutional review board approval (2023-10-002 BC). Informed consent was exempted due to the retrospective nature of the data collection and the use of deidentified CT images. The CT images included a variety of common neurology conditions affecting the skull, brain parenchyma, nasal sinuses, and the eye, and were collected by radiologists who routinely obtain CT images and write image reports based on the images and the patient's medical records. Since Alzheimer's Disease is a progressive degenerative condition predominantly seen in the elderly, the dataset includes images of normal brains, past infarcts that still show manifestations, chronic brain conditions, and acute brain lesions.

### Clinical Visual Instruction Tuning (CVIT)

To address the domain constraints of standard MLLM, we conducted multiple end-to-end visual instruction tuning processes on the multi-image mode Otter foundation model, enhancing its applicability to brain CT radiology features.[25,26] Based on the Flamingo model, the Otter paradigm connects the LLaMA-7B language encoder and the frozen CLIP ViT-L/14 vision encoder via a trainable perceiver resampler module and multiple cross-attention layers inserted into the LLaMA-7B architecture. Within the original LLaMA-7B structure, all modules except for the input/output embeddings were frozen to reduce training costs. The training duration for each resulting model was 12 hours on two NVIDIA A100 GPUs, achieving 3 epochs.

To facilitate multi-image in-context learning capacity, we formulated the data into image-instruction-answer triplets, with the instruction tokenized and the image enhanced prior to input into the model. We designed four distinct fine-tuning conditions including regular visual instruction tuning (RVIT, Plain Instruction, and In-context example Instruction) and clinical visual instruction tuning (CVIT, Template Instruction, and Keyword Instruction), each corresponding to an adherence hierarchy to clinical essence. We named the final instruction-tuned models BrainGPT-plain, BrainGPT-example, BrainGPT-template, and BrainGPT-keyword.

For each visual instruction tuning process, BrainGPT-plain was fine-tuned using plain instruction, conveying the model's role as a radiology assistant; BrainGPT-example was fine-tuned using in-context example instruction, adopting a 3-shot example approach due to available RAM constraints as based on the work of Singhal et al.;[27] BrainGPT-template was fine-tuned using template instruction, involving a structured and predefined set of questions or points that need to be addressed; BrainGPT-keyword was fine-tuned using keyword instruction, focusing on essential areas or categorical guidelines that direct the model's response generation process. Detailed instruction examples can be referenced in Extended Data Fig. 4.

### Dataset preparation

**Training Dataset.** Since the Otter architecture requires image-text pair instances to be of the same size (24 slices), we sampled 365,928 slices from 15,238 scans representing 7,747 patients from a total of 597,335 slices from 15,247 scans representing 7,751 patients for the training process. The system was then tested on 87,312 slices sampled from 145,166 slices of 3,638 CT scans representing 1,938 patients.

**External Validation Dataset.** The CQ500 dataset, consisting of 1,154 available CT scans from 491 patients, was downloaded from the Qure.ai website.[39] The dataset focuses on image features such as brain parenchyma (plain scans), bone (bone scans), and blood vessels (post-contrast scans). Only non-contrast CT scans with slice numbers between 23 and 40 were selected to build the external validation dataset (n=133). This ensures that slice thickness and details are similar to our training dataset and fit in the Otter framework. Ground truth was based on a read.csv file from the CQ500 dataset, and the majority rule was applied among the three raters to summarize the "Mass effect," "Hemorrhage event," and "Midline shift" labels.

### Feature-Oriented Radiology Task Evaluation (FORTE)

We proposed the Feature-Oriented Radiology Task Evaluation (FORTE) framework to capture radiology keywords in terms of Degree (size and intensity), Landmark (location and anatomy), Feature (disease traits and primary findings), and Impression (final diagnosis) of the disease. (For details on the keyword list, see Extended Data Table 7). This list is not just a compilation of keywords but also includes synonyms, allowing the system to recognize a broader array of related terms that may appear in different contexts, addressing the challenge of lexical variability inherent in clinical reports. The F1 score is calculated for each category, providing a multi-faceted evaluation of the system's performance. Additionally, we compute Pearson's correlation coefficient (Pearson's r) for each FORTE category with traditional metrics, offering a deeper understanding of their applicability and limitations in radiological report evaluation.

### Traditional metrics

We compared the clinical evaluation fitness of FORTE against the standard similarity-based evaluation metrics, including BLEU (Bilingual Evaluation Understudy, set range 0-100), METEOR (Metric for Evaluation of Translation with Explicit ORdering, set range 0-100), ROUGE-L (Recall-Oriented Understudy for Gisting Evaluation with Longest Common Subsequence, set range 0-100), and CIDEr-R (Robust Consensus-based Image Description Evaluation, set range 0-1000) (Extended Data Table 8)[19-22] All standard evaluations were executed using the Microsoft Common Objects in Context (MS-COCO) toolkit.[40]

Additionally, to address the list-by-list structure, diverse paraphrasing, and differential diagnosis-oriented negation description of brain CT reports, we incorporated sentence pairing (inspired by cosine similarity calculation in

BERTScore)[41] and negation removal before applying the evaluation formulas. By these means, sentence pairing releases the sequential constraints of disease descriptors, and negation removal reduces the false positives in evaluation reports. Specifically, the reports were embedded and vectorized using the all-mpnet-base-v2 model from the SentenceTransformer library before pairing and scoring.

**Linguistic-embedded *Turing* test**

To examine whether the BrainGPT CT report recapitulates the linguistic texture of radiologist reports, we conducted a *Turing* test involving physicians and radiologists. Each participant was asked to distinguish BrainGPT reports from radiologist reports. The study was structured around four measuring axes:

(1) *Turing* test: Can physicians tell the difference between BrainGPT reports and radiologist reports?

(2) Confidence rate: How confident are the reviewers in their ratings?

(3) Inter-leaved dependency: Do physicians alter their assessments and confidence rates after reviewing the original CT scans?

(4) Linguistic criteria: What is the linguistic rationale behind physicians' impressions?[28]

To explore the aforementioned questions, we collected survey and semi-structured rationale interview data. The physician survey was composed of six caption pairs, each comprising a BrainGPT report and a radiologist report. These examples included diverse disease instances including lacunar infarct, subdural hemorrhage, brain atrophy, and midline (cerebral falx) shift, thereby encompassing a range of both acute and chronic cerebral alterations for expert evaluation. (Details can be referred to Extended Data Fig. 5)

## Code and data availability

All code used for experiments in this study can be found in a GitHub repository (https://github.com/charlierabea/FORTE), which also contains a model weight link to our best instruction-tuned model (BrainGPT-keyword). This study used data from Taipei Veterans General Hospital (TPEVGH) for the training process and CQ500 from Qure.ai for external validation.[39] Data from TPEVGH cannot be released due to IRB regulations, but researchers can access CQ500 via the provided reference. Note that any further distribution of datasets is subject to the terms of use and data-sharing agreements stipulated by the original creators.

## Acknowledgements

Funded by grants of the Taiwan National Science and Technology Council (NSTC 112-2321-B-A49-007, NSTC 1 11-2320-B-A49-028-MY3, NSTC 112-2124-M-038-001, and NSTC 112-2314-B-032-001) and Taipei Veterans G eneral Hospital (V112C-026 and 112VACS-007). This study is based in part on data from the Big Data Center, Taipei Veterans General Hospital (BDC, TPEVGH). The interpretations and conclusions contained herein do not represent the position of Taipei Veterans General Hospital. We would like to thank Professor Albert C. Yang of National Yang Ming Chiao Tung University for the technical resources provided.

## Author Contributions

C.Y.L collected data, designed visual instruction tuning instructions, ran experiments, analyzed results, created figures and wrote the manuscript. K.J.C proposed the FORTE framework, analyzed results, and wrote the manuscript. C.F.Y assisted in model fine-tuning and designed sentence pairing. H.Y.W conducted qualitative analysis and created figures. W.T.C assisted in model fine-tuning and provided technical advice. H.B. assisted in model survey and academic writing. L.C, Y.P.Y, Y.C.C, and S.P.C advised on study design and provided meaningful feedback. J.F.L directed the Turing Test and provided feedback on instruction design and the FORTE framework as a radiologist. K.W.C and S.H.C guided the project and provided critical suggestions for overall direction. All authors contributed to the interpretation and provided critical feedback on the analyses and manuscript.

## Competing Interests

The authors declare no competing interests.

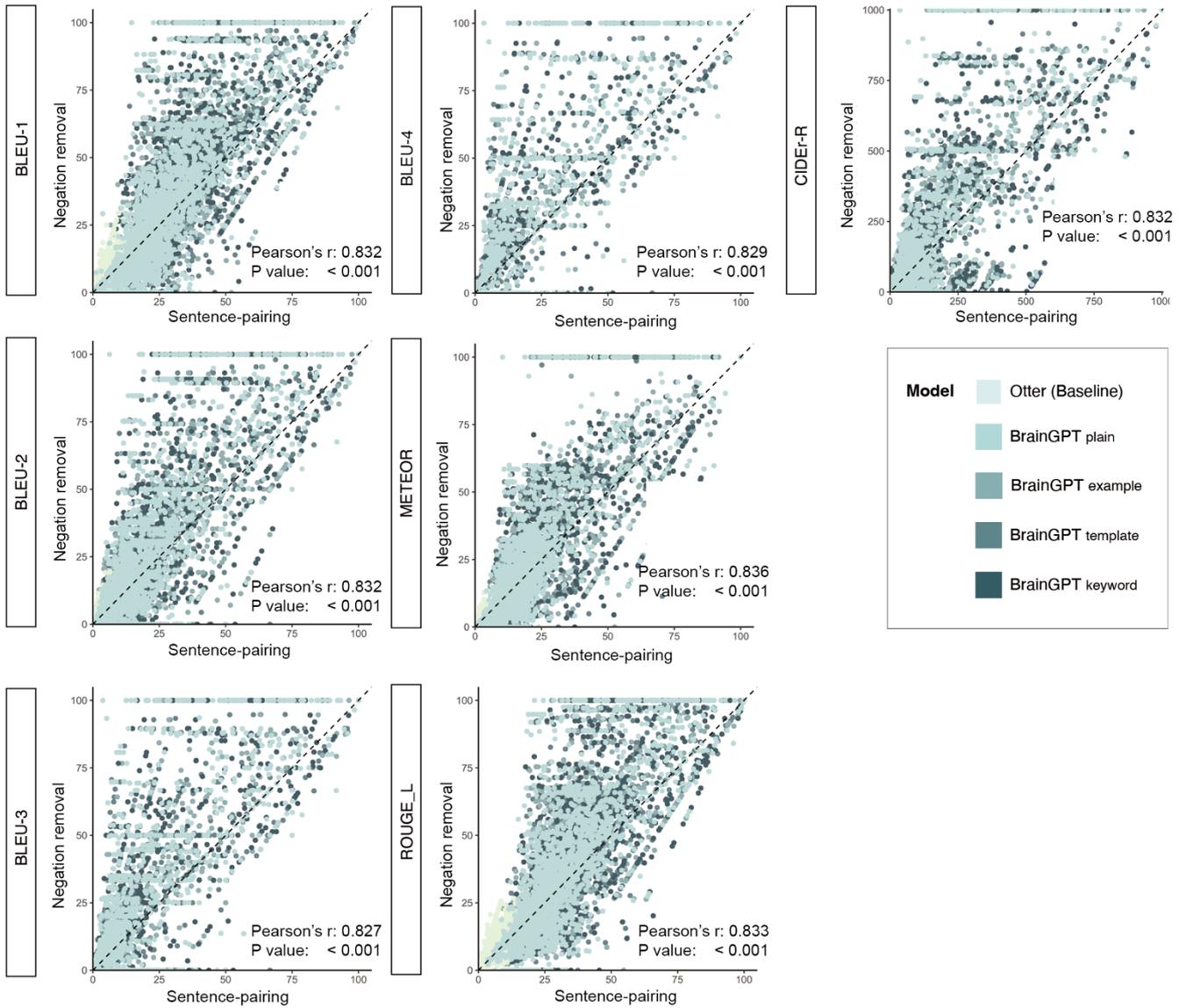

**Extended Data Fig. 1 | Effect of negation removal.** The dashed line denotes equivalence, and each data point corresponds with a text-image instance. The Pearson correlation coefficient (r) and *p* value are indicated on each plot, demonstrating a significant linear correlation with statistical significance in all metrics.

| Modality | Report | Degree | Landmark | Feature | Impression |
|---|---|---|---|---|---|
| Chest X-ray 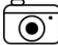 | Findings:<br>...<br>Heart is mildly enlarged.<br>...<br>...<br>Impression<br>Cardiomegaly, consistent with mild congestive heart failure. | Mild | Heart | Enlargement | Cardiomegaly |
| | | Mild | Heart | | Congestive heart failure (CHF) |
| Pathology Image 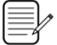 | Findings:<br>...<br>Full thickness invasion of the iris stroma by melanoma cells.<br>...<br>Impression<br>Iris Melanoma. | Full | Iris Stroma | Invasion | Iris Melanoma |
| Retinal Image 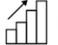 | Findings:<br>...<br>Drusen shadow on choroid plexus<br>...<br>Impression<br>Non-exudative Age-related macular degeneration (AMD) | | Choroid Plexus | Drusen | Non-exudative AMD |

**Extended Data Fig. 2 | Generalization of the Feature-Oriented Radiology Task Evaluation (FORTE).** Based on the semantic information concept, including the degree (quality), landmark (location), feature (primary finding), and impression (disease summary), the FORTE category concept can fit into any medical report generation evaluation framework.

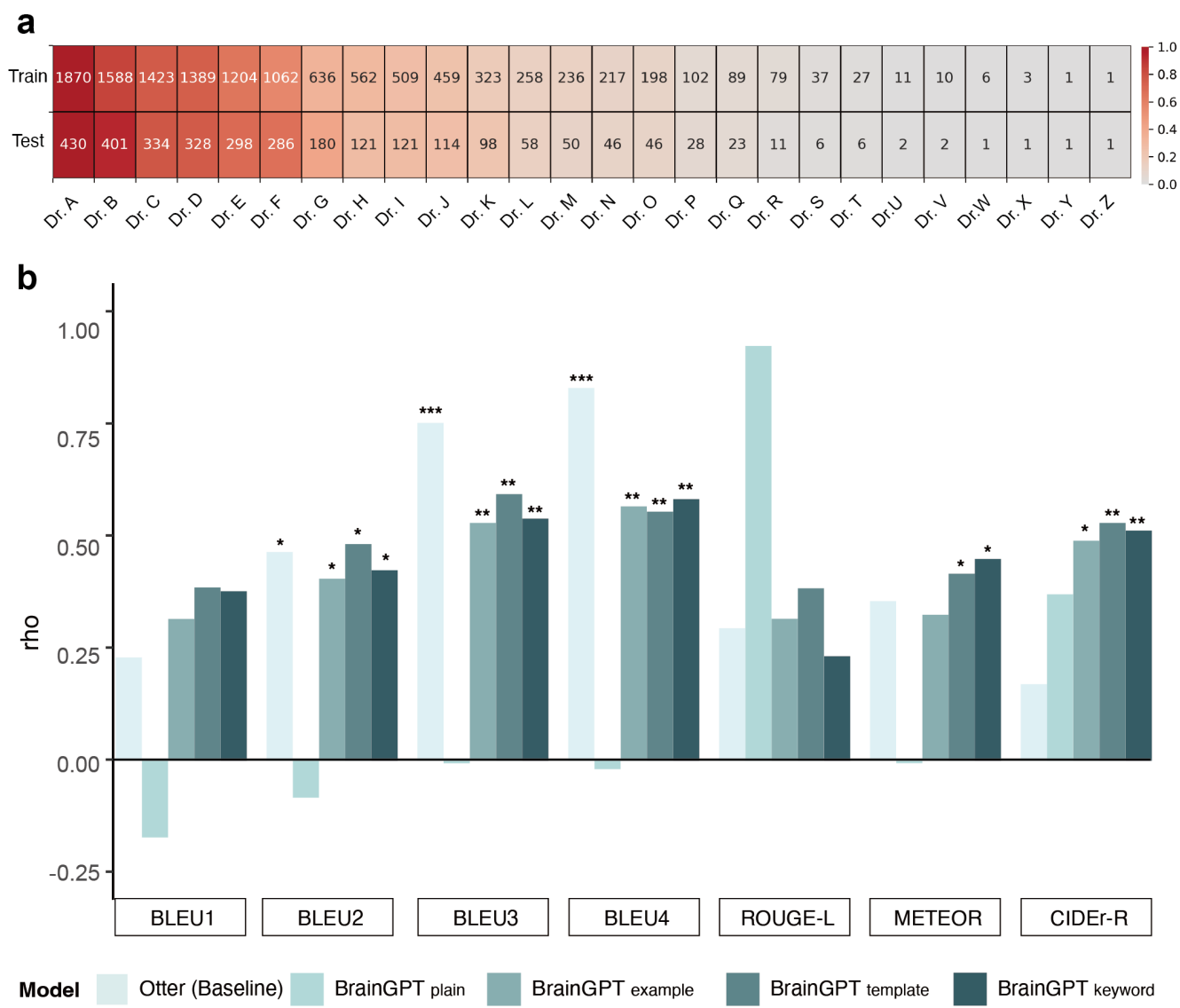

**Extended Data Fig. 3 | The input case imbalance influences the caption writing style. (a)** The number of radiologist inputs in the training and testing portion within the 3D-BrainCT dataset. **(b)** The Spearman correlation analysis between traditional metric scores and the number of radiologist inputs. (* *p* <0.05, ** *p* <0.01, *** *p* <0.001)

| | Plain Instruction |
|---|---|
| You are an AI assistant specialized in radiology topics.<br><br>You are provided with brain CT slices from a single study. The number of slices is 24.<br>Please generate medical descriptions based on the images. | |

| | In-context example Instruction |
|---|---|
| You are an AI assistant specialized in radiology topics.<br>You are provided with brain CT slices from a single study. The number of slices is 24.<br>Please generate medical descriptions based on the images in a consistent style.<br><br><Impression: arteriosclerotic encephalopathy><br>Findings<br>> low density change in the periventricular white matter, most likely as subcortical arteriosclerotic encephalopathy<br>> no intracranial hemorrhage.<br>> normal appearance of insular cortex and no definite effacement of cerebral cortex. no ct evidence of acute infarction of brain.<br>> no ventricular dilatation nor midline shift.<br>> no space-occupying lesion in the brain parenchyma.<br>> bilateral paranasal sinuses and mastoid air cells are well pneumatized.<br>> skull bones appear intact without space-occupying lesion.<br>Conclusion<br>1. low density change in the periventricular white matter, most likely as subcortical arteriosclerotic encephalopathy<br>2. no ct evidence of acute infarction of brain.<br><br><Impression: lacunar infarcts><br>Findings<br>> low density change in the periventricular white matter, most likely as subcortical arteriosclerotic encephalopathy<br>> a few hypodense lesions at bilateral basal ganglia, more in favor of old lacunar infarcts.<br>> no intracranial hemorrhage.<br>> no ventricular dilatation nor midline shift.<br>> no space-occupying lesion in the brain parenchyma.<br>> bilateral paranasal sinuses and mastoid air cells are well pneumatized.<br>> skull bones appear intact without space-occupying lesion.<br>Conclusion<br>several old lacunar infarcts; mild white matter change.<br><br><Impression: SDH(subdural hepatoma/midshift)><br>Findings:<br>> sdh noted at left cerebral convexity, right frontal base, along tentorium cerebelli and cerebral falx, up to 2.0cm.<br>> mass effect with midline shift to right side, about 1cm, and compression of left lateral ventricle, slightly in regression as compared with previous study.<br>> no left side uncal hernation.<br>> bilateral paranasal sinuses and mastoid air cells are well pneumatized.<br>> skull bones appear intact without space-occupying lesion.<br>Conclusion<br>> sdh noted at left cerebral convexity, right frontal base, along tentorium cerebelli and cerebral falx, up to 2.0cm.<br>> mass effect with midline shift to right side, about 1cm, and compression of left lateral ventricle, in regressive change. | |

| | Template Instruction |
|---|---|
| You are an AI assistant specialized in radiology topics.<br><br>You are provided with brain CT slices from a single study. The number of slices is 24.<br>Please generate medical descriptions based on the images in a consistent style.<br><br><Style template><br>Findings:<br>> <is there a midline shift or hemorrhage(intracranial hepatoma/epidural hepatoma/subdural hepatoma)?><br>> <is there a change in ventricular and sulci system?><br>> <Is there a white matter lesion?(lacunar infarction/cortical infarction/subcortical infarction)><br>> <Is the brain parenchyma healthy?(tissue loss(atrophy)/tissue swelling)><br>> <Is there abnormality in high density area?(meningioma/fracture/calcified plaque/arachnoid cyst)><br>> <(Use your domain knowledge)Is there any other abnormality?(herniation/arteriosclerotic encephalopathy/encephalomalacia/wall calcification of cavernous ICA/Air-fluid level)><br><br>Conclusion:<br>1. <summarizing important finding 1><br>2. <summarizing important finding 2> | |

| | Keyword Instruction |
|---|---|
| You are an AI assistant specialized in radiology topics.<br><br>You are provided with brain CT slices from a single study. The number of slices is 24.<br>Please generate medical descriptions based on the images in a consistent style.<br><br>Use the following guidelines:<br>- Degree: Indicate the intensity or state (e.g., normal, mild, chronic, old, etc).<br>- Landmark: Specify the area of interest (e.g., intracerebral, midline, parenchyma, sulci, etc).<br>- Feature: Describe any observed abnormalities (e.g., hemorrhage, atrophy, infarcts, etc).<br>- Impression: Conclude with a clinical impression (e.g., arteriosclerotic encephalopathy, intracerebral hemorrhage, dementia, etc).<br><br>Ensure consistency and clarity in the report. | |

**Extended Data Fig. 4 | Visual instruction tuning instruction design of BrainGPT.** We used instructions and examples from a panel of qualified clinicians to design four distinct instructions for visually tuning BrainGPT from Otter. The resulting models are BrainGPT-plain, BrainGPT-example, BrainGPT-template, and BrainGPT-keyword, with additional parameters aligned with the medical domain.

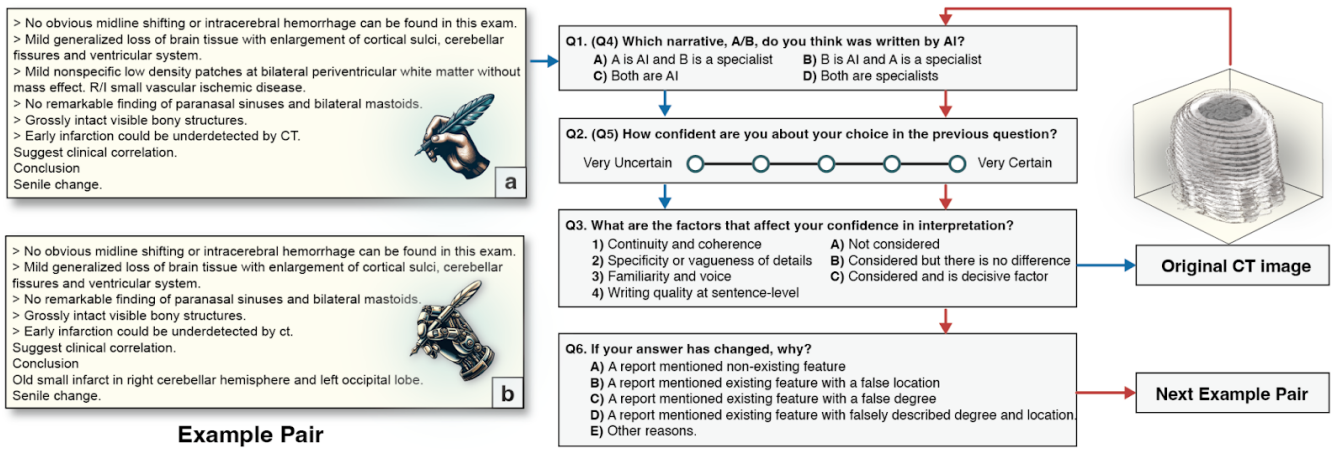

**Extended Data Fig. 5 | The design of our linguistic-embedded *Turing* test.** We utilized a combination of survey and semi-structured interview data. The physician survey consisted of 7 examples (42 questions) and took approximately 30 minutes to complete. Physicians were first asked to choose from four options (the Turing test), followed by rating their confidence level on a 5-point Likert scale and then identifying and rating the importance of four rationales. After being shown the corresponding CT image to the report examples, participants reassessed their choices and confidence levels.

**Extended Data Table 1 | BrainGPT Report Examples.** Shown are examples used in the *Turing* test, with physician reviewer accuracy ranging from 0.09 to 0.36. The traditional metric scores for these high-quality reports were listed.

| Radiologist Ground Truth | BrainGPT Generated Report | Report level | Sentence pairing | Negation removal | *Turing* Test Human expert success rate |
|---|---|---|---|---|---|
| > Mild generalized loss of brain tissue with enlargement of cortical sulci and ventricular system.<br>> White matter low density change is noted in the bilateral periventricular region, mostly likely angiopathic leukoencephalopathy from arteriosclerotic change.<br>> Suspicious lacunar infarction at the left anterior basal ganglia.<br>> *No* intracerebral hemorrhage and no midline shift is seen in the current examination.<br>> Otherwise nothing remarkable. | > Generalized atrophic change of the brain parenchyma with widening of sulci and dilated ventricular system.<br>> *No* obvious midline shifting or intracerebral hemorrhage can be found in this exam.<br>> Old cerebrovascular accident with Encephalomalacia in the right cerebellar hemisphere.<br>> Several lacunar infarcts in the right lentiform nucleus. | BLEU-1: 37.01<br>BLEU-2: 20.19<br>BLEU-3: 8.68<br>BLEU-4: 0<br>METEOR: 18.16<br>ROUGE-L: 29.48<br>CIDEr: 1.49 | BLEU-1: 38.99<br>BLEU-2: 16.78<br>BLEU-3: 0<br>BLEU-4: 0<br>METEOR: 21.42<br>ROUGE-L: 38.85<br>CIDEr: 53.49 | BLEU-1: 46.16<br>BLEU-2: 14.43<br>BLEU-3: 0<br>BLEU-4: 0<br>METEOR: 21.94<br>ROUGE-L: 48.09<br>CIDEr: 71.59 | Before reading CT scan:<br>**0.18**<br><br>After reading CT scan:<br>**0.27** |
| > *No* intracerebral hemorrhage, subdural hematoma, epidural hematoma, subarachnoid hemorrhage or midline shift seen in the current examination.<br>> Tiny low density lesions in the anterior limb of right side internal capsule and corona radiata, compatible with old lacunar infarctions.<br>> Generalized loss of brain tissue with enlargement of cortical sulci and ventricular system.<br>> Low density change is noted in the periventricular white matter of bilateral cerebral hemispheres, mostly likely angiopathic leukoencephalopathy from arteriosclerotic change.<br>> Calcification in the wall of bilateral intracranial vertebral arteries and the cavernous part of bilateral icas.<br>Conclusion<br>Aging process and atrophic change of the brain with old lacunar infarctions in right corona radiata and anterior limb of right internal capsule; no ich, sah, sdh or edh. | > *No* intracerebral hemorrhage, subdural hematoma, epidural hematoma, subarachnoid hemorrhage or midline shift seen in the current examination.<br>> Generalized loss of brain tissue with enlargement of cortical sulci and ventricular system.<br>> Extensive low density change is noted in the periventricular white matter of bilateral cerebral hemispheres, mostly likely angiopathic leukoencephalopathy from arteriosclerotic change.<br>> Tiny low density lesions in bilateral lentiform nuclei, thalami and left corona radiata, compatible with lacunar infarctions.<br>> Calcification in the wall of bilateral intracranial vertebral arteries and the cavernous part of bilateral internal carotid arteries.<br>Conclusion<br>Aging process and atrophic change of the brain with lacunar infarctions in bilateral lentiform nuclei, thalami and left corona radiata; for more detailed information, please see the descriptions above. | BLEU-1: 82.11<br>BLEU-2: 77.83<br>BLEU-3: 73.71<br>BLEU-4: 69.78<br>METEOR: 45.07<br>ROUGE-L: 69.34<br>CIDEr: 376.31 | BLEU-1: 81.84<br>BLEU-2: 78.92<br>BLEU-3: 76.49<br>BLEU-4: 74.46<br>METEOR: 63.4<br>ROUGE-L: 83.63<br>CIDEr: 693.81 | BLEU-1: 78.21<br>BLEU-2: 74.7<br>BLEU-3: 71.78<br>BLEU-4: 69.35<br>METEOR: 56.09<br>ROUGE-L: 80.36<br>CIDEr: 632.57 | Before reading CT scan:<br>**0.27**<br><br>After reading CT scan:<br>**0.36** |
| > *No* obvious midline shifting or intracerebral hemorrhage can be found in this exam.<br>> Mild generalized loss of brain tissue with enlargement of cortical sulci, cerebellar fissures and ventricular system.<br>> *No* remarkable finding of paranasal sinuses and bilateral mastoids.<br>> Grossly intact visible bony structures.<br>> Early infarction could be underdetected by ct.<br>Suggest clinical correlation.<br>Conclusion<br>Old small infarct in right cerebellar hemisphere and left occipital lobe.<br>Senile change. | > *No* obvious midline shifting or intracerebral hemorrhage can be found in this exam.<br>> Mild generalized loss of brain tissue with enlargement of cortical sulci, cerebellar fissures and ventricular system.<br>> Mild nonspecific low density patches at bilateral periventricular white matter without mass effect. R/I small vascular ischemic disease.<br>> *No* remarkable finding of paranasal sinuses and bilateral mastoids.<br>> Grossly intact visible bony structures.<br>> Early infarction could be underdetected by CT. Suggest clinical correlation.<br>Conclusion<br>Senile change. | BLEU-1: 77.5<br>BLEU-2: 76.08<br>BLEU-3: 74.62<br>BLEU-4: 73.47<br>METEOR: 54.52<br>ROUGE-L: 81.03<br>CIDEr: 238.3 | BLEU-1: 74.07<br>BLEU-2: 73.08<br>BLEU-3: 66.67<br>BLEU-4: 55.91<br>METEOR: 71.6<br>ROUGE-L: 75.89<br>CIDEr: 681.33 | BLEU-1: 66.67<br>BLEU-2: 65.39<br>BLEU-3: 57.14<br>BLEU-4: 43.31<br>METEOR: 63.49<br>ROUGE-L: 69.00<br>CIDEr: 590.28 | Before reading CT scan:<br>**0.18**<br><br>After reading CT scan:<br>**0.36** |

**Extended Data Table 1 (continued) | Example: Effect of sentence pairing and negation removal.**

| Radiologist Ground Truth | BrainGPT Generated Report | Report level | Sentence pairing | Negation removal | Turing Test Human expert success rate |
|---|---|---|---|---|---|
| > Nonspecific low density patches in bilateral periventricular white matter without mass effect. r/i small vascular ischemic disease.<br>> Mild generalized enlargement of cortical sulci, cerebellar fissures and ventricles, in favor of mild generalized brain atrophy.<br>> <mark>No</mark> skull bone fracture.<br>Conclusions<br>SDH at left fronto-temporal about 10mm in thickness. | > Presence of thin-layered SDH at right occipital region about 2.8 mm in thickness.<br>> Low dense subdural effusion at left fronto-parietal region, with mass effect to left lateral ventricle and minimal midline shift.<br>> Old lacunar infarction at right thalamus.<br>> Mild generalized loss of brain tissue with enlargement of cortical sulci, cerebellar fissures and ventricular system.<br>> Calcification at the wall of left intracranial vertebral artery and the cavernous part of bilateral internal carotid arteries.<br><mark>No</mark> remarkable finding of paranasal sinuses and bilateral mastoids.<br>> Grossly intact visible bony structures.<br>Impression<br>Low dense subdural effusion at left fronto-parietal region, with mass effect to left lateral ventricle and minimal midline shift. | BLEU-1: 22.12<br>BLEU-2: 16.02<br>BLEU-3: 11.16<br>BLEU-4: 8.43<br>METEOR: 19.45<br>ROUGE-L: 25.06<br>CIDEr: 0 | BLEU-1: 20.43<br>BLEU-2: 11.49<br>BLEU-3: 5.46<br>BLEU-4: 4.69<br>METEOR: 10.59<br>ROUGE-L: 23.1<br>CIDEr: 58.96 | BLEU-1: 23.88<br>BLEU-2: 14.77<br>BLEU-3: 7.01<br>BLEU-4: 6.04<br>METEOR: 12.36<br>ROUGE-L: 27.06<br>CIDEr: 75.8 | Before reading CT scan:<br>**0.18**<br><br>After reading CT scan:<br>**0.36** |
| Findings<br>> Subacute sdh at left frontoparietal convexity, about 19mm in maximal thickness. it caused buckling of left cerebral hemisphere, and midline shift to right side, 4mm.<br>> Several old infarcts at right frontal, parietal, right anterior corona radiata, and bilateral cerebellums.<br>> Atherosclerotic change of bilateral carotid siphons.<br>> Skull bones appear intact without space-occupying lesion.<br>Conclusion<br>1.Subacute sdh at left frontoparietal convexity, about 19mm in maximal thickness.<br>it Caused buckling of left cerebral hemisphere, and midline shift to right side, 4mm.<br>2.Several old infarcts. | > Generalized atrophic change of the brain parenchyma with widening of sulci and dilated ventricular system.<br>> Chronic subdural hematoma about 25mm in thickness in left frontoparietal region with mass effect, post-operative change with bone defect.<br>> Slight midline shifting to the right side. | BLEU-1: 14.21<br>BLEU-2: 6.03<br>BLEU-3: 0<br>BLEU-4: 0<br>METEOR: 7.91<br>ROUGE-L: 17.15<br>CIDEr: 0 | BLEU-1: 13.71<br>BLEU-2: 4.21<br>BLEU-3: 0<br>BLEU-4: 0<br>METEOR: 8.57<br>ROUGE-L: 17.9<br>CIDEr: 14.41 | BLEU-1: 13.71<br>BLEU-2: 4.21<br>BLEU-3: 0<br>BLEU-4: 0<br>METEOR: 8.57<br>ROUGE-L: 17.9<br>CIDEr: 14.41 | After reading CT scan:<br>**0.36**<br><br>After reading CT scan:<br>**0.36** |
| Findings<br>> Moderate amount of subacute-chronic sdh along bilateral f-p-t convexities, about 13mm in maximal thickness. this leads to buckling of bilateral cerebral hemispheres<br>> A small old insult at right frontal white matter.<br>> A few hypodense lesions at bilateral basal ganglia, more in favor of old lacunar infarcts.<br>> <mark>No</mark> midline shift nor brain herniation.<br>> Atherosclerotic change of bilateral carotid siphons.<br>> <mark>No</mark> hydrocephalus.<br>Conclusion<br>Moderate amount of subacute-chronic sdh along bilateral f-p-t convexities, about 13mm in maximal thickness. this leads to buckling of bilateral cerebral hemispheres early infarction could be underdetected by ct. suggest clinical correlation. | > Post-operative change with bone defects in left frontotemporoparietal skull, tissue loss in left anterior temporal, frontal pole and frontal opercular region.<br>> Chronic SDH with linear calcification at left high frontoparietal convexity about 1.0cm in thickness and left high frontotemporal convexity about 0.6cm in thickness.<br>> Minimal SDH noted at right low frontal convexity about 0.3cm in thickness.<br>> <mark>No</mark> midline shift or brain herniation at current study.<br>> Mild generalized loss of brain tissue with enlargement of cortical sulci, cerebellar fissures and ventricular system.<br>> Mild nonspecific low density patches at bilateral periventricular white matter without mass effect. R/I small vascular ischemic disease.<br>> <mark>No</mark> remarkable finding of paranasal sinuses and bilateral mastoids.<br>> Recommend clinical correlation and follow up. | BLEU-1: 28.8<br>BLEU-2: 14.45<br>BLEU-3: 6.98<br>BLEU-4: 4.09<br>METEOR: 16.70<br>ROUGE-L: 22.68<br>CIDEr: 0 | BLEU-1: 17.27<br>BLEU-2: 8.40<br>BLEU-3: 3.38<br>BLEU-4: 2.6<br>METEOR: 13.6<br>ROUGE-L: 22.59<br>CIDEr: 42.21 | BLEU-1: 12.72<br>BLEU-2: 3.43<br>BLEU-3: 0<br>BLEU-4: 0<br>METEOR: 10.08<br>ROUGE-L: 17.38<br>CIDEr: 25.42 | After reading CT scan:<br>**0.09**<br><br>After reading CT scan:<br>**0.27** |

**Extended Data Table 2 | Results of the traditional evaluation metrics.**

| Model Performance | BLEU-1 | BLEU-2 | BLEU-3 | BLEU-4 | METEOR | ROUGE-L | CIDEr-R |
|---|---|---|---|---|---|---|---|
| **Report-level Evaluation** | | | | | | | |
| Otter (baseline) | 8.73 | 2.82 | 0.33 | 0 | 3.79 | 10.18 | 5.9 |
| BrainGPT plain | 31.59 | 21.91 | 16.12 | 12.95 | 19.4 | 29.52 | 19.1 |
| BrainGPT example | 27.9 | 19.15 | 13.79 | 11.08 | 17.11 | 28.5 | 21.28 |
| BrainGPT template | 28.14 | 19.64 | 14.52 | 11.85 | 17.94 | 29.34 | 31.47 |
| BrainGPT keyword | 28.55 | 20.16 | 15.08 | 12.37 | 17.96 | 29.45 | 31.26 |
| **Sentence pairing** | | | | | | | |
| Otter (baseline) | 7.94 | 1.54 | 0.13 | 0 | 3.46 | 8.41 | 4.34 |
| BrainGPT plain | 30.12 | 20.23 | 13.54 | 11.49 | 21.28 | 33.03 | 125.86 |
| BrainGPT example | 32.29 | 21.39 | 12.99 | 11.5 | 22.66 | 35 | 132.38 |
| BrainGPT template | 34.69 | 23.36 | 15.04 | 13.19 | 24.64 | 37.2 | 147.92 |
| BrainGPT keyword | 34.96 | 23.85 | 15.55 | 13.59 | 24.95 | 37.46 | 153.3 |
| **Negation removal** | | | | | | | |
| Otter (baseline) | 6.87 | 1.03 | 0.15 | 0.01 | 3.02 | 7.88 | 4.12 |
| BrainGPT plain | 38.01 | 26.8 | 20.04 | 17.89 | 26.84 | 41.59 | 189.52 |
| BrainGPT example | 43.68 | 30.41 | 20.99 | 19.47 | 29.72 | 47.17 | 206.22 |
| BrainGPT template | 44.64 | 31.24 | 22.29 | 20.32 | 30.02 | 47.77 | 209.9 |
| BrainGPT keyword | 44.35 | 31.3 | 22.45 | 20.38 | 30.13 | 47.6 | 211.77 |

**Extended Data Table 3 | Two-tailed Mann-Whitney U test results of the traditional metrics.** Whole report level evaluation revealed that all metrics improved in BrainGPT compared to the general Otter model, with the Mann-Whitney U test showing very strong statistical significance (p < 0.001) for BLEU-1, BLEU-2, BLEU-3, BLEU-4, METEOR, and ROUGE-L. However, none of the quantitative metrics showed significant improvements across the fine-tuned models (p > 0.05). The sentence pairing method showed strong statistical significance in BLEU-1, METEOR, and ROUGE when comparing BrainGPT-plain with other instruction-tuned models (p < 0.01). Compared with BrainGPT-example, both BrainGPT-template and BrainGPT-keyword showed significant improvement in all metrics (p < 0.05). However, no significant difference was observed between the BrainGPT-template and the BrainGPT-keyword, suggesting a similar effect of clinical instructions.

| Model Comparison | BLEU-1 | BLEU-2 | BLEU-3 | BLEU-4 | METEOR | ROUGE-L | CIDEr-R |
|---|---|---|---|---|---|---|---|
| **Whole report level Evaluation** | | | | | | | |
| Otter vs. BrainGPT plain | <0.001*** | <0.001*** | <0.001*** | <0.001*** | <0.001*** | <0.001*** | <0.001*** |
| Otter vs. BrainGPT example | <0.001*** | <0.001*** | <0.001*** | <0.001*** | <0.001*** | <0.001*** | 0.045* |
| Otter vs. BrainGPT template | <0.001*** | <0.001*** | <0.001*** | <0.001*** | <0.001*** | <0.001*** | <0.001*** |
| Otter vs. BrainGPT keyword | <0.001*** | <0.001*** | <0.001*** | <0.001*** | <0.001*** | <0.001*** | 0.008** |
| BrainGPT plain vs. BrainGPT example | 0.081 | 0.099 | 0.122 | 0.131 | 0.123 | 0.279 | 0.198 |
| BrainGPT plain vs. BrainGPT template | 0.14 | 0.253 | 0.421 | 0.461 | 0.254 | 0.545 | 0.687 |
| BrainGPT plain vs. BrainGPT keyword | 0.144 | 0.305 | 0.528 | 0.532 | 0.221 | 0.653 | 0.941 |
| BrainGPT example vs. BrainGPT template | 0.926 | 0.868 | 0.673 | 0.54 | 0.877 | 0.742 | 0.505 |
| BrainGPT example vs. BrainGPT keyword | 0.719 | 0.708 | 0.564 | 0.396 | 0.492 | 0.426 | 0.407 |
| BrainGPT template vs. BrainGPT keyword | 1 | 0.912 | 0.911 | 0.88 | 0.625 | 0.863 | 0.933 |
| **Sentence pairing** | | | | | | | |
| Otter vs. BrainGPT plain | <0.001*** | <0.001*** | <0.001*** | <0.001*** | <0.001*** | <0.001*** | <0.001*** |
| Otter vs. BrainGPT example | <0.001*** | <0.001*** | <0.001*** | <0.001*** | <0.001*** | <0.001*** | <0.001*** |
| Otter vs. BrainGPT template | <0.001*** | <0.001*** | <0.001*** | <0.001*** | <0.001*** | <0.001*** | <0.001*** |
| Otter vs. BrainGPT keyword | <0.001*** | <0.001*** | <0.001*** | <0.001*** | <0.001*** | <0.001*** | <0.001*** |
| BrainGPT plain vs. BrainGPT example | <0.001*** | 0.194 | <0.001*** | <0.001*** | 0.006** | <0.001*** | 0.017* |
| BrainGPT plain vs. BrainGPT template | <0.001*** | 0.001** | 0.047* | 0.001** | <0.001*** | <0.001*** | 0.821 |
| BrainGPT plain vs. BrainGPT keyword | <0.001*** | <0.001*** | 0.769 | 0.144 | <0.001*** | <0.001*** | 0.17 |
| BrainGPT example vs. BrainGPT template | 0.015* | 0.049* | <0.001*** | 0.001** | 0.037* | 0.015* | 0.022* |
| BrainGPT example vs. BrainGPT keyword | 0.008** | 0.004** | <0.001*** | <0.001*** | 0.016* | 0.005** | <0.001*** |
| BrainGPT-template vs. BrainGPT keyword | 0.835 | 0.352 | 0.12 | 0.114 | 0.757 | 0.694 | 0.125 |

**Extended Data Table 4 | Results of the Feature-Oriented Radiology Task Evaluation (FORTE).**

| Instruction | F1 Degree | F1 Landmark | F1 Feature | F1 Impression | FORTE Average |
|---|---|---|---|---|---|
| **All outputs** | | | | | |
| BrainGPT plain | 0.498 | 0.444 | 0.484 | 0.467 | 0.474 |
| BrainGPT example | 0.529 | 0.491 | 0.536 | 0.578 | 0.534 |
| BrainGPT template | 0.54 | 0.534 | 0.578 | 0.64 | 0.573 |
| BrainGPT keyword | 0.548 | 0.533 | 0.574 | 0.649 | 0.576 |
| **Negation removal** | | | | | |
| BrainGPT plain | 0.604 | 0.653 | 0.65 | 0.751 | 0.665 |
| BrainGPT example | 0.654 | 0.712 | 0.688 | 0.698 | 0.688 |
| BrainGPT template | 0.671 | 0.708 | 0.697 | 0.758 | 0.708 |
| BrainGPT keyword | 0.661 | 0.706 | 0.693 | 0.779 | 0.71 |

**Extended Data Table 5 | Two-tailed Mann-Whitney U test results of the Feature-Oriented Radiology Task Evaluation (FORTE).** FORTE revealed that all categories improved in BrainGPT compared to the general Otter model, with very strong statistical significance ($p < 0.001$) in the Mann-Whitney U test. Comparing BrainGPT-plain with other instruction-tuned models, FORTE showed statistical significance ($p < 0.001$) in every keyword category. Compared with BrainGPT-example, both the BrainGPT-template and BrainGPT-keyword showed significant improvement in degree, landmark, and impression ($p < 0.05$). However, no significant difference was observed between BrainGPT-template and BrainGPT-keyword, suggesting a similar effect of clinical visual instruction tuning instructions.

| Model Comparison | F1 Degree | F1 Landmark | F1 Feature | F1 Impression | FORTE Average |
|---|---|---|---|---|---|
| **All outputs** | | | | | |
| Otter vs. BrainGPT plain | <0.001*** | <0.001*** | <0.001*** | <0.001*** | <0.001*** |
| Otter vs. BrainGPT example | <0.001*** | <0.001*** | <0.001*** | <0.001*** | <0.001*** |
| Otter vs. BrainGPT template | <0.001*** | <0.001*** | <0.001*** | <0.001*** | <0.001*** |
| Otter vs. BrainGPT keyword | <0.001*** | <0.001*** | <0.001*** | <0.001*** | <0.001*** |
| BrainGPT plain vs. BrainGPT example | <0.001*** | <0.001*** | <0.001*** | <0.001*** | <0.001*** |
| BrainGPT plain vs. BrainGPT template | <0.001*** | <0.001*** | <0.001*** | <0.001*** | <0.001*** |
| BrainGPT plain vs. BrainGPT keyword | <0.001*** | <0.001*** | <0.001*** | <0.001*** | <0.001*** |
| BrainGPT example vs. BrainGPT template | 0.001** | <0.001*** | 0.173 | 0.044* | <0.001*** |
| BrainGPT example vs. BrainGPT keyword | 0.001** | <0.001*** | 0.913 | 0.031* | <0.001*** |
| BrainGPT template vs. BrainGPT keyword | 0.82 | 0.342 | 0.186 | 0.996 | 0.867 |
| **Negation removal** | | | | | |
| Otter vs. BrainGPT plain | 0.03* | 0.921 | 0.085 | N/A | <0.001*** |
| Otter vs. BrainGPT example | 0.008** | <0.001*** | 0.116 | N/A | <0.001*** |
| Otter vs. BrainGPT template | 0.006** | <0.001*** | 0.062 | N/A | <0.001*** |
| Otter vs. BrainGPT keyword | 0.013* | <0.001*** | 0.115 | N/A | <0.001*** |
| BrainGPT plain vs. BrainGPT example | <0.001*** | <0.001*** | <0.001*** | 0.155 | <0.001*** |
| BrainGPT plain vs. BrainGPT template | <0.001*** | <0.001*** | <0.001*** | 0.515 | <0.001*** |
| BrainGPT plain vs. BrainGPT keyword | <0.001*** | <0.001*** | <0.001*** | 0.694 | <0.001*** |
| BrainGPT example vs. BrainGPT template | 0.761 | 0.045* | 0.291 | 0.477 | 0.074 |
| BrainGPT example vs. BrainGPT keyword | 0.946 | <0.001*** | 0.908 | 0.271 | 0.179 |
| BrainGPT template vs. BrainGPT keyword | 0.152 | 0.071 | 0.288 | 0.933 | 0.6858 |

**Extended Data Table 6 | SOTA level performance of BrainGPT.** Despite a smaller model size (7B), the BrainGPT models achieved better results in the 3D-BrainCT dataset than the performance of Med-PaLM (84B) in the relatively simple MIMIC Chest X-ray dataset.[16,26,27]

| Model | Dataset | BLEU-1 | BLEU-4 | ROUGE-L |
|---|---|---|---|---|
| Previous SOTA | MIMIC Chest X-ray (MIMIC-CXR) | 39.48 | 13.3 | 29.6 |
| PaLM-E (84B) | MIMIC Chest X-ray (MIMIC-CXR) | 19.86 | 4.6 | 16.53 |
| Med-PaLM M (84B) | MIMIC Chest X-ray (MIMIC-CXR) | 32.31 | 11.31 | 27.29 |
| Baseline CT2Rep with CT-Net (Customized Transformer) | CT-RATE (Chest CT) | 44.3 | 35.4 | 44.2 |
| CT2Rep (Customized Transformer) | CT-RATE (Chest CT) | 46 | 36.9 | 45.9 |
| Otter (7B) | 3D-BrainCT | 6.87 | 0.01 | 7.88 |
| BrainGPT plain (7B) | 3D-BrainCT | 38.01 | 17.89 | 41.59 |
| BrainGPT example (7B) | 3D-BrainCT | 43.68 | 19.47 | 47.17 |
| BrainGPT template (7B) | 3D-BrainCT | 44.64 | 20.32 | 47.77 |
| BrainGPT keyword (7B) | 3D-BrainCT | 44.35 | 20.38 | 47.6 |

**Extended Data Table 7 | Feature-Oriented Radiology Task Evaluation (FORTE) Keyword Categories and List.** This list is not merely a compilation of keywords but also includes synonyms, enabling the system to recognize a broader array of related terms that may appear in different contexts. For each identified keyword, the system retrieves its primary representative or returns the keyword itself if no such representative exists. This approach ensures that semantically similar keywords are treated as equivalent, addressing the challenge of lexical variability inherent in clinical reports.

| Degree | | | | | |
|---|---|---|---|---|---|
| No | Moderate | Normal/ Intact/ Healthy | | Focal/ Territorial/ Localized | |
| Regressive | Chronic | Inapparent/ Nonspecific | | Obvious/ Apparent | |
| Slight/ Subtle/ Faint/ Minimal/ Tiny/ Minute/ Mild/ Low/ Thin | | Generalized/ Extensive/ Grossly/ Obviously/ Clearly/ Prominent/ Marked | | Abnormal/ Atypical/ Irregular/ Anomalous | |
| **Landmark** | | | | | |
| Posterior | Anterior | Ventral | Lateral | Frontal | Occipital |
| Frontoparietal | Occipitoparietal | Parasagittal | Periventricular | Paramedian | Supraclinoid |
| Epidural | Subdural | Subarachnoid | Intracerebral | Subcortical | Supratentorial |
| Ethmoid | Sphenoid | Paranasal | Maxillary | | |
| Gray-white matter | Corpus callosum | White matter/ Lacunar | Ventricular/ Ventricle | Corona radiata | External capsule |
| Genu | Thalamus/ Thalami | Basal ganglia | Striatum | Globus pallidus | Putamen |
| Caudate nucleus/ Caudate nuclei/ Caudate | | Lentiform nuclei/ Lentiform nucleus | | | |
| Parenchyma | Insular/ Insula | Lobes/ Lobe | Centrum semiovale | Venous sinus | Mastoids/ Mastoid |
| Vertex Region | Vascular/ Artery | Vertebral artery | MCA/ Middle cerebral artery | ICA/ Internal carotid arteries/ Internal carotid siphons | |
| Segment | Scalp | Nasopharynx | Mucus/ Mucosal | Lens/ Eyes/ Eyelid | Foramen magnum |
| Sulci/ Sylvian fissure/ Fissure/ Cerebral fissure / Sulcus | | Hemispheres/ Hemisphere/ Ipsilateral | | Bone/ Skull | Midline/ Falx |
| **Feature** | | | | | |
| Shift | Retention | Congestion | Air-fluid level | Thickening | CSF/ Cerebrospinal fluid |
| Density | Crowding | Buckeling | Wedge shape | Lipoma | Hyper dense/ Hyperdensity |
| Metastasis | Infarcts/ Ischemic | Hemorrhage/ Hematoma | Patches/ hypointensity | Dilated/ Widening/ Enlarged/ Enlargement | |
| Perifocal edema/ Swelling | | Hypodensities/ Low density/ Attenuation/ Hypodensity | | Calcified plaque/ Calcification/ Arteriosclerosis/ Calcified/ Calcification/ Calcified-like | |
| Cerebrovascular/ Angiopathic | | Ground-glass opacity/ opacity | | Atrophy/ Atrophic/ Degeneration | |
| Space occupying lesion/ Mass effect | | Effusion/ Fluid Accumulation | | Pneumatized/ Air cell/ pneumocephalus | |
| **Impression** | | | | | |
| Herniation | Hydrocephalus | Sinusitis | Pneumocephalus | Encephalomalacia | Leptomeningeal seeding |
| Recommend | Craniotomy | Meningitis | Leukoencephalopathy | Cateract | Communicating hydrocephalus |
| Meningioma | Thrombosis | Arachnoid cyst | Shunting Catheter | Brain Atrophy | Transependymal effusions |
| Regressive change | Brain Metastasis | Fibrous dysplasia | Wallerian Degeneration | Mastoid effusion | Aging/Aging change/ Senile change |
| SDH/ Subdural hematoma | EDH/ Epidural hematoma | IVH/ Intraventricular hemorrhage | | Aging/Aging change/ Senile change | |

**Extended Data Table 8 | Comparison among traditional evaluation metrics and the FORTE.**

| Evaluation Metric | Method | Purpose |
|---|---|---|
| **BLEU** (Bilingual Evaluation Understudy) | Quantification of n-gram co-occurrence frequencies (unigram = single word) | Machine Translation |
| **METEOR** (Metric for Evaluation of Translation with Explicit ORdering) | Word-to-word matching with precision, recall, and penalty calculations | Machine Translation |
| **ROUGE-L** (Recall-Oriented Understudy for Gisting Evaluation with Longest Common Subsequence) | Longest Common Subsequence (LCS) | Machine Translation/ Document Summarization |
| **CIDEr-R** (Robust Consensus-based Image Description Evaluation) | Cosine similarity of the frequency-inverse document frequency (TF-IDF) vectors | General Image Captioning |
| **FORTE** (Feature-Oriented Radiology Task Evaluation) | F1-score of clinical word-to-word matching in four categories "Degree", "Landmark", "Feature", and "Impression." | Radiology Image Report Generation |